\documentclass{article} 
\PassOptionsToPackage{table}{xcolor}
\usepackage{workborne_wechat}

\renewcommand{\workbornedate}{2026-10-10}

\usepackage{amsmath,amsfonts,bm}

\def\eqref#1{equation~\ref{#1}}

\def\1{\bm{1}}

\DeclareMathAlphabet{\mathsfit}{\encodingdefault}{\sfdefault}{m}{sl}
\SetMathAlphabet{\mathsfit}{bold}{\encodingdefault}{\sfdefault}{bx}{n}

\usepackage{booktabs}
\usepackage{tabularx}
\usepackage{array}
\usepackage{amssymb}
\usepackage{graphicx}
\usepackage{multirow}
\usepackage{pifont}
\usepackage[table]{xcolor}
\usepackage{colortbl}
\usepackage{float}
\usepackage{wrapfig}
\usepackage{placeins}
\usepackage{caption}
\usepackage{url}
\usepackage{hyperref}

\newcommand{\gain}[1]{\textcolor{WBGreen}{\scriptsize\,(+#1)}}

\usepackage{booktabs}
\usepackage{graphicx}
\usepackage[table]{xcolor}
\usepackage{multirow}
\usepackage{makecell}

\definecolor{sectiongreen}{RGB}{235,247,238}

\newcommand{\modelicon}[1]{%
  \raisebox{-0.12em}{%
    \makebox[\dimexpr1em+2\fboxsep+2\fboxrule\relax][c]{%
      \rule[-0.28em]{0pt}{1.56em}%
      \raisebox{-0.06em}{#1}%
    }%
  }%
}

\newcommand{\best}[1]{\textbf{#1}}

\definecolor{WBGreen}{HTML}{2D6A4F}
\definecolor{WBDeep}{HTML}{1B4332}
\definecolor{WBMid}{HTML}{74A98C}
\definecolor{WBLight}{HTML}{EAF4EE}
\definecolor{WBHighlight}{HTML}{D8EDE0}
\definecolor{WBZebra}{HTML}{F5FAF7}
\definecolor{WBGroup}{HTML}{EDF5F0}
\definecolor{WBMuted}{HTML}{66756C}

\newcolumntype{Y}{>{\raggedright\arraybackslash}X}
\newcolumntype{C}{>{\centering\arraybackslash}X}

\usepackage[most]{tcolorbox}
\usepackage{enumitem}
\definecolor{WBPromptFrame}{HTML}{2C3E50}
\definecolor{WBPromptAccent}{HTML}{2B807B}
\definecolor{WBPromptRule}{HTML}{D7E0E7}
\newtcolorbox{wbpromptbox}[1]{
  enhanced, breakable, lines before break=5,
  colback=white, colbacktitle=white,
  colframe=WBPromptFrame, coltitle=WBPromptFrame,
  boxrule=0.6pt, arc=3pt,
  borderline west={3pt}{0pt}{WBPromptAccent},
  title={#1}, title after break={#1 (continued)},
  fonttitle=\sffamily\bfseries\normalsize,
  fontupper=\rmfamily\normalsize,
  titlerule=0.35pt, titlerule style={WBPromptRule},
  left=10pt, right=10pt, top=7pt, bottom=7pt,
  toptitle=6pt, bottomtitle=6pt,
  before skip=\baselineskip, after skip=6pt,
  before upper={\raggedright\setlength{\parindent}{0pt}\setlength{\parskip}{2pt}},
  pad at break*=5pt
}
\newcommand{\wbpromptheading}[1]{\par\addvspace{2pt}\noindent\textbf{#1}\par\nobreak}
\newcommand{\wbpromptvariable}[1]{\textcolor{WBPromptAccent}{#1}}
\newcommand{\wbpromptomission}[1]{\par\nobreak\noindent\textcolor{WBMuted}{#1}\par}
\newlist{wbpromptitems}{itemize}{1}
\setlist[wbpromptitems]{label=\textbullet,leftmargin=1.25em,itemsep=0.5pt,parsep=0pt,topsep=2pt,beginpenalty=10000}
\newlist{wbpromptsteps}{enumerate}{1}
\setlist[wbpromptsteps]{label=\arabic*.,leftmargin=1.5em,itemsep=0.5pt,parsep=0pt,topsep=2pt,beginpenalty=10000}

\title{\textcolor{WBTemplateGreen}{WorkBorne:} Discovering Agentic Tasks from\\
Real-World Artifact Transitions}

\author{%
Junyi Zhou\textsuperscript{1,2}, Yufei Wang\textsuperscript{2,\dag}, Qiyuan Zhang\textsuperscript{1,2}, Chuhan Wu\textsuperscript{2}, Zhengxi Li\textsuperscript{3},\\
Wentao Zhao\textsuperscript{2}, Jie Zhou\textsuperscript{2}, Le Tian\textsuperscript{2}, Xiao Zhou\textsuperscript{2}, Chen Ma\textsuperscript{1,\dag}\\[0.7em]
{\normalfont\WBDisplayFont\fontsize{10}{12}\selectfont \textsuperscript{1}City University of Hong Kong \qquad \textsuperscript{2}WeChat AI, Tencent}\\[0.2em]
{\normalfont\WBDisplayFont\fontsize{10}{12}\selectfont \textsuperscript{3}MBZUAI}\qquad
{\normalfont\rmfamily\fontsize{10}{12}\selectfont \textsuperscript{\dag}Corresponding authors}
}

\hypersetup{
  pdftitle={WorkBorne: Discovering Agentic Tasks from Real-World Artifact Transitions},
  pdfauthor={Junyi Zhou, Yufei Wang, Qiyuan Zhang, Chuhan Wu, Zhengxi Li, Wentao Zhao, Jie Zhou, Le Tian, Xiao Zhou, Chen Ma}
}

\begin{document}
\raggedbottom

\maketitle
\begingroup
\makeatletter
\let\@footnotetext\H@@footnotetext
\long\def\@makefntext#1{\noindent#1}
\makeatother
\renewcommand{\thefootnote}{}
\footnotetext{\fontsize{8}{10}\selectfont\raggedright\hypersetup{urlcolor=black}%
Junyi Zhou \href{mailto:junyizhou8-c@my.cityu.edu.hk}{\texttt{<junyizhou8-c@my.cityu.edu.hk>}},
Yufei Wang \href{mailto:garyyfwang@tencent.com}{\texttt{<garyyfwang@tencent.com>}},
Chen Ma \href{mailto:chenma@cityu.edu.hk}{\texttt{<chenma@cityu.edu.hk>}}}
\endgroup


\begin{abstract}
Developing terminal-based agents for long-horizon execution demands synthetic tasks that tightly couple complex instructions with authentic, interactive environments. However, current automated synthesis suffers from a fundamental asymmetry: \emph{instruction-oriented} methods fabricate low-fidelity, synthetic environments, whereas \emph{resource-oriented} methods treat real artifacts merely as static inputs, degrading task instructions into shallow, trivial operations. To address this limitation, we introduce \textbf{WorkBorne}, a synthesis framework that uses \emph{artifact transitions} as the central primitive for modeling the full life-cycle of agentic tasks. In \textbf{WorkBorne}, each transition specifies who performs the task, which artifacts anchor the initial state and intended outcome, and the corresponding relationship between them. 
Specifically, WorkBorne first constructs coherent and diverse artifact corpora from iterative web retrieval, then discovers non-trivial transitions by identifying meaningful input-output artifact assemblies and the relations that connect them. Each discovered transition is finally instantiated into grounded agentic tasks. We construct 1,061 artifact corpora and 7,882 task environments, yielding 15,485 high-quality trajectories, with 53.6\% exceeding 30 assistant turns. Experiments on Qwen3.5-27B and Qwen3.5-35B-A3B show that \textbf{WorkBorne} substantially improves over base models (i.e., improving Terminal-Bench 2.1 from 43.1\% to 61.8\% and ClawEval General score from 75.0 to 80.2) and achieves the strongest overall performance among other synthesis baselines (e.g., TMax~\citep{ivison2026tmaxsimplerecipeterminal}, RST~\citep{li2026recursive}, Terminal-Universe~\citep{wu2026terminaluniverseturningagenttrajectories}). Further analysis shows that transition is strongly associated with execution depth, while corpus breadth and transition depth provide complementary scaling gains. 
\end{abstract}

\section{Introduction}
\label{sec:introduction}

As terminal-based agents tackle increasingly complex professional tasks~\citep{merrill2026terminalbenchbenchmarkingagentshard,ye2026clawevaltrustworthyevaluationautonomous}, training and evaluating them heavily rely on high-quality trajectories that capture sustained, long-horizon interaction~\citep{yang2026makes,DBLP:journals/corr/abs-2602-01244}.
Unlike conversational data, such trajectories emerge from agents executing task instructions within complex interactive environments~\citep{cheng2026terminalworld,shi2026facet}.
However, tasks that genuinely induce long-horizon execution remain exceptionally rare~\citep{li2026recursive,li2026longhorizonterminalbenchtestinglimitsagents}.
Most daily tasks are inherently shallow, and human instructions are often designed to minimize interaction steps~\citep{han2026ridewaybenchmarkingefficienttask}, leaving limited opportunities to elicit iterative reasoning, execution, and verification. 
Therefore, the central challenge in agentic data synthesis is to discover tasks whose instructions and environments jointly induce sustained interaction. 


Existing automated synthesis methods generally fail to produce long-horizon tasks due to an instruction-environment asymmetry in their design.
\textbf{Instruction-oriented methods} begin with a generated task instruction and use language models to reconstruct an environment~\citep{zhu2026termigen,cheng2026terminalworld}. Although this guarantees clear instruction framing, synthesized artifacts often fail to capture the richness, coherence, and incidental complexity of real-world workspaces~\citep{shi2026facet,tang2026workspacebench10benchmarkingai}. \textbf{Resource-oriented methods} instead discover tasks from APIs~\citep{DBLP:conf/nips/LiuHZZLKTYLFNYS24}, web resources~\citep{liu2025webexplorerexploreevolvetraining}, or tool interactions~\citep{dong2026agentworldscalingrealworldenvironment,wu2026terminaluniverseturningagenttrajectories}. These methods provide authentic workspaces, but treat the artifacts merely as passive task input, making it easy to generate shallow instructions like reading a file and producing a brief summary~\citep{hua2026cliuniverse}. 
In short, this asymmetry prevents existing paradigms from combining authentic workspace richness with instruction complexity to drive sustained agent interaction.

In this work, we observe that authentic agentic workflows naturally follow a simple \textbf{3W} primitive: \emph{who} works on \emph{what input} toward \emph{what output}. 
In real-world professional environments, both the initial inputs and the intended outputs are grounded in concrete artifacts (i.e., specific files or digital items), while the substantive task lies in the transition connecting them. 
This perspective motivates us to treat \textbf{artifact transitions} as a unified abstraction to jointly model task instructions and environments, and we then propose \textbf{WorkBorne}, a fully automated framework that discovers agentic tasks from real-world artifact transitions. 
Specifically, WorkBorne first constructs coherent and diverse artifact corpora through iterative web retrieval, providing rich materials from authentic workflows. 
To discover non-trivial transitions, WorkBorne selects personas for each corpus, uses them to assemble grounded input-output artifact anchors, and then infers the required transition by reasoning how the inputs must be adapted, extended, or integrated to realize the anchored output. Finally, each validated transition is instantiated into aligned task instructions, executable environments, and evaluation rubrics, naturally ensuring them remain grounded in the same workspace. 

Using WorkBorne, we construct 1,061 artifact corpora and 7,882 executable tasks, yielding 15,485 high-quality trajectories across 14 agent harnesses, with 53.6\% exceeding 30 turns. 
Fine-tuning Qwen3.5-27B and Qwen3.5-35B-A3B~\citep{qwen3.5} on these trajectories (yeilding \textbf{WorkBorne-27B} and \textbf{WorkBorne-35B-A3B}) achieves substantial gains across benchmarks, with WorkBorne-27B boosting Terminal-Bench 2.1~\citep{merrill2026terminalbenchbenchmarkingagentshard} accuracy from 43.1\% to 61.8\% and ClawEval General~\citep{ye2026clawevaltrustworthyevaluationautonomous} score from 75.0 to 80.2.
A controlled construction ablation (\S\ref{sec:construction_ablation}) shows that transition modeling increases average execution length from 27.8 to 44.5 assistant turns over an input-only pipeline, while improving Terminal-Bench 2.1 from 48.3\% to 51.7\%. 
Building on this, an empirical investigation into structural execution factors (\S\ref{sec:partition_execution}) shows that execution depth depends little on raw artifact quantity ($\eta=0.015$), but strongly on transition structure. Input-output artifact assembly shows a much stronger association ($\eta=0.314$), with output-anchor form more informative than input form ($\eta=0.306$ vs.\ $0.183$). 
Finally, scaling studies (\S\ref{sec:breadth_depth_scaling}) further show that corpus breadth and transition depth provide complementary gains, while broader corpus coverage consistently performs better under matched training budgets.

\section{Related Work}
\label{sec:related_work}

\subsection{Terminal-Based Agents.}
Terminal-based agents have evolved from interactive code execution~\citep{yang2023intercode} to repository-scale software engineering~\citep{jimenez2024swebench,wang2024codeact,yang2024sweagent,wang2025openhands}, and more recently toward open-ended, general-purpose command-line interaction~\citep{merrill2026terminalbenchbenchmarkingagentshard,chen2026tuabench}.
This trajectory has sparked parallel efforts to construct training environments from real codebases~\citep{pan2025training} and to optimize execution efficiency~\citep{ren2026self,mavali2026moreless}.
As agent development rapidly accelerates toward sustained, long-horizon execution~\citep{desai2026swemarathon,yang2026t1}, the primary bottleneck has pivoted from agent architecture design to scalable data synthesis: specifically, generating task suites that scale in volume without compromising the structural complexity and execution fidelity these complex tasks demand.

\subsection{Terminal Task Synthesis.}
Existing task synthesis methodologies broadly fall into two complementary paradigms, each incurring a fundamental quality trade-off when scaled up.
1) \textbf{Instruction-oriented methods} adopt a top-down route, initiating synthesis from task specifications and constructing matching environments~\citep{zhu2026termigen,gandhi2026endless,ivison2026tmaxsimplerecipeterminal,zhao2026nex}, with subsequent extensions escalating complexity via compositional skill graphs~\citep{fan2026scalable,cheng2026terminalworld}, recursive task expansion~\citep{li2026recursive,pan2026metatask}, and adversarial calibration~\citep{meng2026calibforge}. While effective at driving textual difficulty, these environments remain LM-fabricated, systematically lacking the \emph{incidental complexity} inherent to authentic workspaces.
2) \textbf{Resource-oriented methods} take a bottom-up route by anchoring synthesis in real-world codebases or pre-existing execution traces~\citep{yang2026swesmith,lin2026cligym,wu2026terminaluniverseturningagenttrajectories}. Although they provide realistic workspace ecosystems, they treat artifacts as static initial conditions, causing synthesized instructions to frequently degrade into shallow inspection operations~\citep{pi2026data,yang2026makes}.
Recent hybrid approaches seek to enforce instruction-environment alignment via executable-state grounding~\citep{shi2026facet}, real-material verification~\citep{hua2026cliuniverse}, and database-level grounding~\citep{wang2026agent}; yet, they continue to treat instructions and environments as decoupled synthesis targets.

\section{WorkBorne}
\label{sec:workborne}

This section presents \textbf{WorkBorne}, a framework that models terminal-based agent tasks through unified artifact transitions.
Below, we firstly formalize agentic tasks as state transitions over real-world artifact spaces (\S\ref{sec:workborne_motivation}), detail the three-stage synthesis pipeline covering \textit{corpus construction}, \textit{transition discovery}, and \textit{task instantiation} (\S\ref{sec:workborne_method}), and conclude with an empirical profile of the generated dataset and execution trajectories (\S\ref{sec:dataset_profile}).

\begin{figure}[t]
    \centering
    \includegraphics[width=\textwidth]{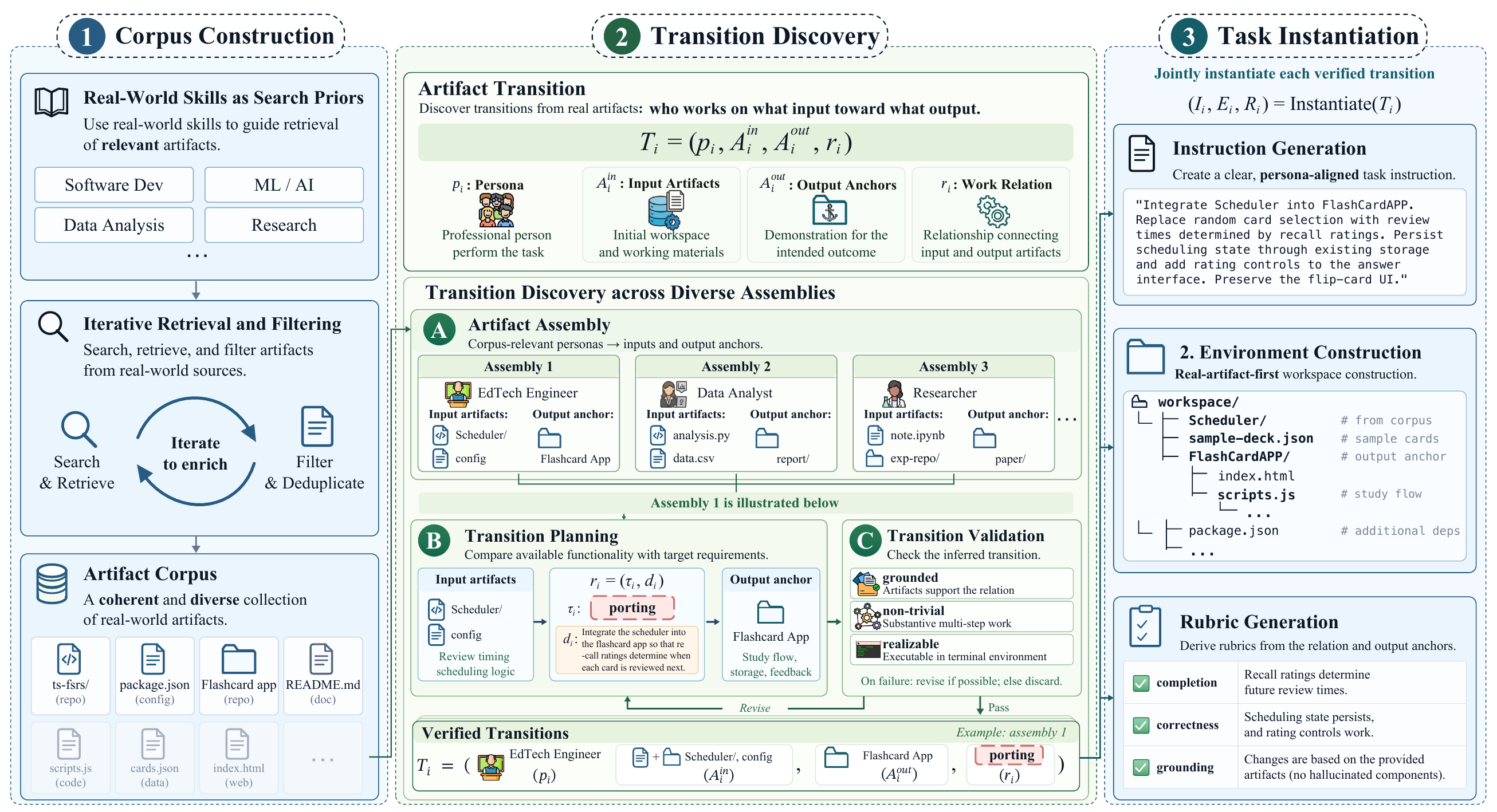}
    \caption{
    \textbf{Overview of WorkBorne.}
    WorkBorne models agentic tasks as artifact transitions $\mathcal{T}=(p,\mathcal{A}^{\mathrm{in}},\mathcal{A}^{\mathrm{out}},r)$ and synthesizes them in three stages:
    \textbf{(1) Artifact Corpus Construction:} building real-world artifact pools;
    \textbf{(2) Transition Discovery:} discovering grounded transitions through assembly, planning, and validation; and
    \textbf{(3) Task Instantiation:} compiling transitions into aligned instructions, executable environments, and evaluation rubrics.
    }
    \label{fig:workborne_overview}
\end{figure}

\subsection{Task Formulation}
\label{sec:workborne_motivation}

A terminal-based agentic task is fundamentally an execution workflow driven by complex task instructions and authentic environments. 
At a high level, such a workflow follows a \textbf{3W} primitive: \textbf{\emph{who}} works on \textbf{\emph{what input}} toward \textbf{\emph{what output}}. 
Real-world artifacts anchor this process by physically defining both the initial workspace and the target outcome.
However, existing synthesis paradigms often model only isolated parts of this workflow, either fabricating environments around generated instructions or treating real artifacts merely as static inputs, resulting in shallow instructions.
To capture the task lifecycle holistically, \textbf{WorkBorne} instead jointly models these \textbf{3W} elements through an \emph{artifact transition} $\mathcal{T}$, formalized as a four-tuple: 

\begin{equation}
\mathcal{T}_i =
\left(
p_i,
\mathcal{A}_i^{\mathrm{in}},
\mathcal{A}_i^{\mathrm{out}},
r_i
\right),
\label{eq:artifact_transition}
\end{equation}

where $p_i$ denotes the persona performing the task, $\mathcal{A}_i^{\mathrm{in}}$ denotes the set of input artifacts that form the initial workspace, $\mathcal{A}_i^{\mathrm{out}}$ denotes a set of output anchors that demonstrate the intended outcome, and $r_i$ specifies the relationship between the input and output artifacts. 
With $\mathcal{T}_i$ serving as the task skeleton, WorkBorne realizes task creation by first discovering valid transitions from raw artifact space and then instantiating them into an aligned instruction, environment, and rubrics (\S\ref{sec:workborne_method}).

\subsection{Transition-Driven Task Synthesis}
\label{sec:workborne_method}


\paragraph{Artifact Corpus Construction.} Synthesizing realistic agent tasks requires workspaces composed of artifacts that naturally co-occur in professional workflows, rather than arbitrary or isolated documents.
To build such cohesive workspaces, WorkBorne uses \emph{agent skills}~\citep{anthropic2025agentskills} as retrieval seeds.
Each skill defines a specific domain context (e.g., data engineering, API integration) and guides a search agent to gather a multi-format file collection, such as source code, raw datasets, and technical documentation, that logically belong to the same execution pipeline. We curate 1,061 high-quality skills from a public pool of over 10,000 skills to cover a wide spectrum of real-world scenarios. For each skill $s_c$, the search agent receives its workflow description, toolchain, and expected deliverables to formulate web queries and retrieve an initial artifacts collection.
Under a fixed retrieval budget, the agent iteratively expands this collection by filtering duplicates and conducting targeted follow-up searches for missing or sparsely covered file formats (e.g., retrieving configuration files to complement a codebase).
The resulting corpus $\mathcal{C}_c=\{A_1,\ldots,A_{N_c}\}$ forms a realistic multi-file workspace, where each $A_i$ represents an individual artifact.
Finally, artifacts in $\mathcal{C}_c$ are summarized into \emph{file cards} $G_c$, which record file paths, metadata, key entities, and contents to provide a structured index for subsequent transition discovery.

\paragraph{Transition Discovery.}
Directly prompting an agent to imagine tasks from arbitrary artifacts often leads to shallow or repetitive operations~\citep{hua2026cliuniverse}.
To discover substantive and diverse transitions, WorkBorne anchors task generation in \emph{artifact transitions} that explicitly bridge concrete input artifacts to target output anchors within each workspace $\mathcal{C}_c$.
Driven by the structured index $G_c$, this discovery process operates in three sequential steps:

\textbf{(1) Artifact Assembly.}
Given an exploration budget $N$, this step forms $N$ artifact assemblies $(p_i,\mathcal{A}_i^{\mathrm{in}},\mathcal{A}_i^{\mathrm{out}})$ from each corpus $\mathcal{C}_c$. 
Specifically, we first randomly sample $2N$ candidate personas from PersonaHub~\citep{ge2025scalingsyntheticdatacreation} and use an agent to filter them down to the $N$ personas most relevant to the domain context of $\mathcal{C}_c$.
For each selected persona $p_i$, an agent inspects the file cards $G_c$ to select two disjoint artifact subsets $\mathcal{A}_i^{\mathrm{in}}, \mathcal{A}_i^{\mathrm{out}} \subset \mathcal{C}_c$.
The input set $\mathcal{A}_i^{\mathrm{in}}$ naturally matches the persona's expertise, while the output anchor set $\mathcal{A}_i^{\mathrm{out}}$ acts as a real-world demonstration of the intended outcome, specifying structural, formatting, or constraints that the final work must satisfy.

\textbf{(2) Transition Planning.}
Given an assembly $(p_i,\mathcal{A}_i^{\mathrm{in}},\mathcal{A}_i^{\mathrm{out}})$, this step infers the underlying transition relation $r_i$ that connects the input artifacts to the target output anchors.
The transition agent analyzes reusable content, functionality, or structure in $\mathcal{A}_i^{\mathrm{in}}$, and reasons how these components must be adapted, extended, or integrated to realize $\mathcal{A}_i^{\mathrm{out}}$.
The resulting relation $r_i$ consists of a open-ended operation tag and a detailed task description specifying the required execution path.
For instance, given a review-scheduling library \emph{ts-fsrs} as $\mathcal{A}_i^{\mathrm{in}}$ and a flashcard study application \emph{SimpleFlashCard} as $\mathcal{A}_i^{\mathrm{out}}$, an educational technology engineer persona plans a non-trivial \emph{porting} transition: adapting the scheduling logic to the application's study flow so that each card is reviewed at an appropriate time. 
Together, $\mathcal{T}_i = (p_i,\mathcal{A}_i^{\mathrm{in}},\mathcal{A}_i^{\mathrm{out}},r_i)$ establishes a complete artifact transition.

\textbf{(3) Transition Validation.}
A dedicated validation agent evaluates whether the artifact assembly genuinely supports the inferred relation $r_i$, verifies that the task demands non-trivial multi-step execution rather than simple file copying, and confirms that it can be realized in a simulated environment.
Transitions failing these criteria are either refined or discarded.

\paragraph{Task Instantiation.}
Once an artifact transition $\mathcal{T}_i$ is validated, WorkBorne jointly instantiates it into the three core components of an executable agent task: a task instruction $\mathcal{I}_i$, an execution environment $\mathcal{E}_i$, and evaluation rubrics $\mathcal{R}_i$:

\begin{equation}
(\mathcal{I}_i,\mathcal{E}_i,\mathcal{R}_i)
=
\operatorname{Instantiate}(\mathcal{T}_i),
\label{eq:task_instantiation}
\end{equation}

where all three components are grounded in the same transition $\mathcal{T}_i$ to guarantee mutual alignment and verifiability. Specifically, three specialized agent modules construct these components as follows: 

\textbf{(1) Instruction generation ($\mathcal{I}_i$).}
An instruction agent synthesizes the persona-aligned prompt $\mathcal{I}_i$ directly from transition $\mathcal{T}_i$.
The persona $p_i$ defines the professional background and framing, while the transition relation $r_i$ together with the artifact sets $(\mathcal{A}_i^{\mathrm{in}}, \mathcal{A}_i^{\mathrm{out}})$ dictates the explicit goal, expected deliverables, and procedural constraints.
Consequently, the generated instruction $\mathcal{I}_i$ faithfully reflects the substantive work required by the transition.

\textbf{(2) Environment construction ($\mathcal{E}_i$).}
Guided by transition $\mathcal{T}_i$ and instruction $\mathcal{I}_i$, an environment agent sets up the execution workspace $\mathcal{E}_i$.
To maximize realism, WorkBorne adopts a \emph{real-artifact-first} policy: input artifacts $\mathcal{A}_i^{\mathrm{in}}$ from corpus $\mathcal{C}_c$ are directly copied to populate the initial workspace, modified only when task-specific adaptations are necessary, and synthesized only as a last resort.
In addition, output anchors $\mathcal{A}_i^{\mathrm{out}}$ are carefully sanitized to retain essential structural context without leaking reference solutions.

\textbf{(3) Evaluation rubric synthesis ($\mathcal{R}_i$).}
A rubric agent derives a structured set of verifiable evaluation rubrics $\mathcal{R}_i$ from transition relation $r_i$ and output anchors $\mathcal{A}_i^{\mathrm{out}}$.
These rubrics establish concrete criteria assessing task completion, functional correctness, and grounding in the provided workspace, ensuring objective and automated trajectory evaluation.

\subsection{Data Collection}
\label{sec:dataset_profile}

\begin{figure}[H]
\centering
\newsavebox{\datasetprofilebox}
\begin{lrbox}{\datasetprofilebox}
\begin{minipage}[t]{0.47\textwidth}
\vspace{0pt}
\centering
\footnotesize
\renewcommand{\arraystretch}{0.90}
\captionof{table}{WorkBorne dataset profile.}
\label{tab:env_profile}
\vspace{0.5mm}
\begin{tabularx}{\linewidth}{Y r}
\toprule
\multicolumn{2}{l}{\textbf{Corpora and tasks}} \\
\midrule
Tasks / corpus & 7,882 / 1,061 \\
Selected source artifacts & 28,712 \\
Median tasks per corpus & 8 \\
\midrule
\multicolumn{2}{l}{\textbf{Environment}} \\
\midrule
Avg.\ files per task & 350 \\
Avg.\ environment size & 30.2\,MB \\
Avg.\ directory depth & 5.5 \\
Avg.\ rubrics per task & 10.4 \\
Corpus-sourced files & 98.8\% \\
\midrule
\multicolumn{2}{l}{\textbf{Trajectory}} \\
\midrule
Harnesses / trajectories & 14 / 15,485 \\
Mean / median assistant turns & 37.5 / 32 \\
Long-horizon rate ($>$30) & 53.6\% \\
\bottomrule
\end{tabularx}
\end{minipage}
\end{lrbox}%
\usebox{\datasetprofilebox}%
\hfill
\begin{minipage}[t][\dimexpr\ht\datasetprofilebox+\dp\datasetprofilebox\relax][s]{0.51\textwidth}
\vspace{0pt}
\centering
\small
\includegraphics[width=\linewidth,height=\dimexpr\ht\datasetprofilebox+\dp\datasetprofilebox-3\baselineskip-12pt\relax,keepaspectratio]{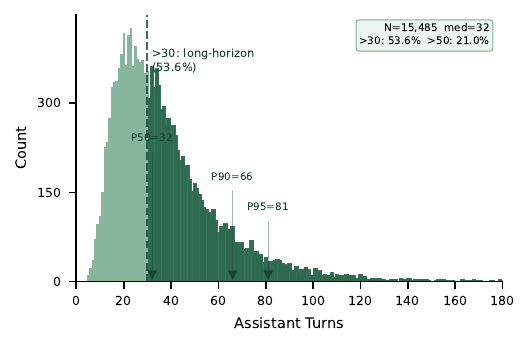}
\vfill
\captionsetup{justification=raggedright,singlelinecheck=false}
\captionof{figure}{Assistant turns in retained trajectories.\\
Darker bars mark those exceeding 30 turns\\
(53.6\%); the tail extends to 353.}
\label{fig:traj_distribution}
\end{minipage}
\end{figure}


We deploy Opus-4.6~\citep{anthropic2026opus46} across the WorkBorne framework to synthesize 1,061 artifact corpora and 7,882 executable task environments.
To collect execution trajectories without overfitting to a single interface or logging schema, we execute rollouts using Kimi-K3~\citep{kimiteam2026kimik3openfrontier} as the teacher model across 14 mainstream agent harnesses (Appendix~\ref{app:harness_details}).
From approximately $24,000$ raw trajectories, we filter out infrastructure failures, normalize heterogeneous harness logs into a unified multi-turn format, and evaluate each rollout against task-specific rubrics using Opus-4.6.
Retaining rollouts with scores of at least 8 out of 10 yields $15,485$ high-quality training trajectories.


As summarized in Table~\ref{tab:env_profile}, the $1,061$ corpora contain $28,712$ selected source artifacts, supporting a median of 8 tasks per corpus.
Each task environment averages 350 files (30.2~MB) and 10.4 rubrics, with 98.8\% of file occurrences copied directly from the underlying source corpora.
Furthermore, the retained trajectories average 37.5 assistant turns with a median of 32 turns, where 53.6\% exceed 30 turns and 21.0\% exceed 50 turns (Figure~\ref{fig:traj_distribution}).
These metrics confirm that transition-driven synthesis reliably induces sustained, long-horizon agent interactions at scale.

\section{Experiments}
\label{sec:experiments}

In this section, we evaluate the performance and generalizability of models trained on Workborne trajectories across three challenging benchmarks: Terminal-Bench Lite, Terminal-Bench 2.1~\citep{merrill2026terminalbenchbenchmarkingagentshard}, and ClawEval~\citep{ye2026clawevaltrustworthyevaluationautonomous}. Our evaluation focuses on three core aspects: performance against existing synthesis methods, generalization across heterogeneous agent frameworks, and the impact of model active capacity.

\subsection{Experimental Setup}
\label{sec:setup}

\paragraph{Training configuration.} We use MS-SWIFT with Megatron to perform multi-turn SFT on Qwen3.5-27B and Qwen3.5-35B-A3B using 128 H20 GPUs. We train on all 15,485 WorkBorne trajectories for three epochs, with a peak learning rate of $4\times10^{-5}$, a global batch size of 32, and a sequence length of 256K tokens with packing. Before training, we check the training data for contamination against Terminal-Bench and ClawEval tasks using 13-gram matching (Appendix~\ref{app:contamination}). Full training settings appear in Appendix~\ref{app:training_configs}.

\paragraph{Evaluation configuration.} We use Terminal-Bench Lite and Terminal-Bench 2.1 to assess challenging, long-horizon task execution. To cover general agent work, we also evaluate the General subset of ClawEval. For both Terminal-Bench benchmarks, we use Terminus2 with temperature 1.0, and a context window of 128K tokens. We report mean raw reward for TB Lite and pass@1 for TB 2.1, both averaged over three runs. For ClawEval, we use the official harness with temperature 0.6 and a context window of 256K tokens. We report pass@3, strict joint accuracy (pass$^\wedge$3), and average score. Full evaluation settings appear in Appendix~\ref{app:evaluation_configs}.

\begin{table*}[t]
\centering
\caption{
\textbf{Main results on three agentic benchmarks:}
Terminal-Bench Lite, Terminal-Bench 2.1, and ClawEval.
We report mean raw reward for TB Lite, pass@1 for TB 2.1, and pass@3, pass\textasciicircum 3, and average score
for ClawEval. Our Terminal-Bench scores are averaged over three runs.
The best result among synthesis methods is shown in \best{bold},
and the second-best result is \underline{underlined}.
Green values in parentheses indicate absolute improvements over the
corresponding Qwen3.5 base model.
\textsuperscript{$\ast$} marks scores we measure on the released models under
the configurations in Appendix~\ref{app:evaluation_configs}; other scores are
taken from the corresponding reports.
}
\label{tab:main_results}

\vspace{2mm}

\resizebox{0.92\textwidth}{!}{%
\renewcommand{\arraystretch}{1.22}%
\setlength{\extrarowheight}{2pt}%
\begin{tabular}{
@{}l
c
cc
ccc@{}
}
\toprule

\multirow{2}{*}{\textbf{Model}}
&
\multirow{2}{*}{\textbf{Size}}
&
\multicolumn{2}{c}{\textbf{Terminal-Bench}}
&
\multicolumn{3}{c}{\textbf{ClawEval (General)}}
\\

\cmidrule(lr){3-4}
\cmidrule(lr){5-7}

&
&
\textbf{TB Lite}
&
\textbf{TB 2.1}
&
\textbf{Pass@3}
&
\textbf{Pass\textasciicircum 3}
&
\textbf{Score}
\\

\midrule

\rowcolor{sectiongreen}
\multicolumn{7}{c}{%
\textbf{\textit{Reference Frontier Models}}
}
\\

\modelicon{\includegraphics[width=1.12em,height=1.12em,keepaspectratio]{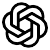}}\ GPT-5.5~\citep{openai2026gpt55}
& --
& --
& 78.0
& -- & -- & --
\\

\modelicon{\includegraphics[width=1.12em,height=1.12em,keepaspectratio]{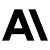}}\ Opus-4.6~\citep{anthropic2026opus46}
& --
& --
& 70.1
& 80.8 & 70.8 & 80.6
\\

\modelicon{\includegraphics[width=1.12em,height=1.12em,keepaspectratio]{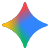}}\ Gemini-3.1-Pro~\citep{googledeepmind2026gemini31pro}
& --
& --
& 65.8
& 80.8 & 55.9 & 76.6
\\

\midrule

\rowcolor{sectiongreen}
\multicolumn{7}{c}{%
\textbf{\textit{Open-source Models}}
}
\\

\modelicon{\includegraphics[width=1.12em,height=1.12em,keepaspectratio]{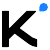}}\ Kimi-K2.6~\citep{moonshot2026kimik26}
& 1T
& --
& 59.9
& 78.9 & 61.5 & 77.6
\\

\modelicon{\includegraphics[width=1.12em,height=1.12em,keepaspectratio]{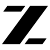}}\ GLM-5~\citep{glm5team2026glm5vibecodingagentic}
& 744B
& 70.6
& 44.9
& 73.9 & 57.1 & 73.8
\\

\modelicon{\includegraphics[width=1.12em,height=1.12em,keepaspectratio]{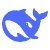}}\ DeepSeek-V4-flash-Preview~\citep{deepseekai2026deepseekv4highlyefficientmilliontoken}
& 284B
& --
& 61.8
& 78.3 & 57.8 & 74.6
\\

\modelicon{\includegraphics[width=1.12em,height=1.12em,keepaspectratio]{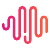}}\ MiniMax-M2.7~\citep{chen2026minimaxm2seriesminiactivations}
& 230B
& --
& 55.4
& 72.0 & 49.7 & 71.8
\\

\midrule

\rowcolor{sectiongreen}
\multicolumn{7}{c}{%
\textbf{\textit{Base \& Synthesis methods}}
}
\\

\modelicon{\includegraphics[width=1.12em,height=1.12em,keepaspectratio]{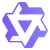}}\ Qwen3.5-35B-A3B~\citep{qwen3.5}
& 35B
& 58.8\textsuperscript{$\ast$}
& 39.7\textsuperscript{$\ast$}
& 77.0\textsuperscript{$\ast$} & 62.1\textsuperscript{$\ast$} & 74.7\textsuperscript{$\ast$}
\\


\modelicon{\includegraphics[width=1.12em,height=1.12em,keepaspectratio]{figures/icon/qwen-color.pdf}}\ Qwen3.5-27B~\citep{qwen3.5}
& 27B
& 68.3\textsuperscript{$\ast$}
& 43.1\textsuperscript{$\ast$}
& 77.6\textsuperscript{$\ast$} & 63.4\textsuperscript{$\ast$} & 75.0\textsuperscript{$\ast$}
\\

\modelicon{\includegraphics[width=1.12em,height=1.12em,keepaspectratio]{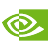}}\ Nemotron-32B~\citep{pi2026data}
& 32B
& 48.6\textsuperscript{$\ast$}
& 25.1\textsuperscript{$\ast$}
& 39.1\textsuperscript{$\ast$} & 4.4\textsuperscript{$\ast$} & 36.0\textsuperscript{$\ast$}
\\

\modelicon{\includegraphics[width=1.12em,height=1.12em,keepaspectratio]{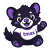}}\ TMax-27B~\citep{ivison2026tmaxsimplerecipeterminal}
& 27B
& 68.6
& 44.9
& \underline{81.4\textsuperscript{$\ast$}} & \best{67.7\textsuperscript{$\ast$}} & \underline{79.6\textsuperscript{$\ast$}}
\\

\modelicon{\includegraphics[width=1.12em,height=1.12em,keepaspectratio]{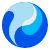}}\ RST-SFT-27B~\citep{li2026recursive}
& 27B
& 71.9\textsuperscript{$\ast$}
& 46.8\textsuperscript{$\ast$}
& 79.5\textsuperscript{$\ast$} & 59.6\textsuperscript{$\ast$} & 75.7\textsuperscript{$\ast$}
\\

\modelicon{\includegraphics[width=1.12em,height=1.12em,keepaspectratio]{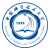}}\ FACET-Terminal-27B~\citep{shi2026facet}
& 27B
& \underline{77.4\textsuperscript{$\ast$}}
& 47.6
& 80.1\textsuperscript{$\ast$} & 57.8\textsuperscript{$\ast$} & 77.1\textsuperscript{$\ast$}
\\

\modelicon{\includegraphics[width=1.12em,height=1.12em,keepaspectratio]{figures/icon/qwen-color.pdf}}\ Terminal-Universe-27B~\citep{wu2026terminaluniverseturningagenttrajectories}
& 27B
& -
& \underline{58.1}
& - & - & -
\\

\midrule

\rowcolor{sectiongreen}
\multicolumn{7}{c}{%
\textbf{\textit{Ours}}
}
\\

\textbf{\modelicon{\includegraphics[width=1.12em,height=1.12em,keepaspectratio]{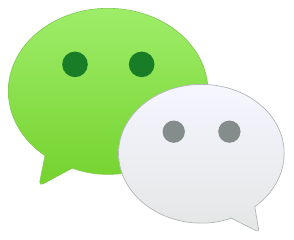}}\ WorkBorne-35B-A3B}
& 35B
& 66.5\gain{7.7}
& 48.3\gain{8.6}
& 80.7\gain{3.7}
& 65.2\gain{3.1}
& 78.3\gain{3.6}
\\

\textbf{\modelicon{\includegraphics[width=1.12em,height=1.12em,keepaspectratio]{figures/logo/wechat.pdf}}\ WorkBorne-27B}
& 27B
& \best{84.3}\gain{16.0}
& \best{61.8}\gain{18.7}
& \best{82.0}\gain{4.4}
& \underline{66.5}\gain{3.1}
& \best{80.2}\gain{5.2}
\\

\bottomrule
\end{tabular}%
}

\end{table*}

\subsection{Main Results}
\label{sec:main_results}

Table~\ref{tab:main_results} presents the results across three benchmarks. We summarize the key findings as follows:

\paragraph{State-of-the-art performance from synthesized supervision.}
WorkBorne establishes top-tier performance among synthesis methods, delivering substantial gains across all benchmarks. WorkBorne-27B reaches 84.3\% on Terminal-Bench Lite and 61.8\% on Terminal-Bench 2.1, improving over its Qwen3.5-27B base by 16.0 and 18.7 percentage points. Using \textbf{only 15K synthesized trajectories}, the 27B model competes directly with substantially larger open models on Terminal-Bench 2.1—matching DeepSeek-V4-flash-Preview (61.8\%) and outperforming Kimi-K2.6 (59.9\%).

\paragraph{Gains persist across heterogeneous agent workflows.}
The performance gains transfer robustly across fundamentally distinct agent architectures and evaluation harnesses. While Terminal-Bench evaluates long-horizon terminal execution via Terminus2, ClawEval measures broader agent capabilities under claw-agent. WorkBorne-27B achieves +4.4\% pass@3 and +5.2\% overall score on ClawEval, alongside a +3.1\% increase in strict joint accuracy (pass$^\wedge$3). The improvement in pass$^\wedge$3 confirms that WorkBorne supervision fosters consistent task completion across repeated trials.

\paragraph{Greater active capacity unlocks larger gains from WorkBorne.}
The value of WorkBorne's long-horizon supervision scales directly with the learner's active capacity. While the MoE variant Qwen3.5-35B-A3B improves steadily across benchmarks (+8.6\% on Terminal-Bench 2.1), the dense Qwen3.5-27B extracts dramatically larger gains from the identical 15K trajectory budget (+18.7\% on TB 2.1 and +5.2\% vs. +3.6\% on ClawEval average). This gap on complex, long-horizon tasks indicates that dense parameters provide the active capacity required to fully absorb rich, multi-stage interaction trajectories.

\section{Analysis and Ablation Studies}
\label{sec:analysis}

In this section, we analyze the core design choices driving WorkBorne's performance gains: \textbf{grounding tasks in real-world artifacts} and \textbf{artifact transition modeling}. We conduct systematic experiments to address three key aspects: 
\textbf{(i) Construction Pipeline Ablation (\S\ref{sec:construction_ablation}):} whether WorkBorne's transition modeling provides additional execution depth and downstream training value beyond baselines; 
\textbf{(ii) Transition Structure and Execution Depth (\S\ref{sec:partition_execution}):} which properties of agentic tasks are associated with deeper execution, and how transition structures shape long-horizon behavior; and 
\textbf{(iii) Corpus and Transition Scaling (\S\ref{sec:breadth_depth_scaling}):} how to scale WorkBorne data efficiently by allocating budgets between corpus breadth and transition depth. 
All comparisons are conducted on Qwen3.5-27B under matched trajectory budgets.

\subsection{Construction Pipeline Ablation}
\label{sec:construction_ablation}

\begin{table}[!htbp]
\centering
\caption{Environment, trajectory, and performance comparison across construction pipelines. Each pipeline contributes 1,000 environments and 1,000 training trajectories. Counts are means over the sampled environments and retained trajectories; long-horizon rate is the percentage of retained trajectories with $>30$ assistant turns.}
\label{tab:ablation_performance}
\footnotesize
\setlength{\tabcolsep}{3pt}
\renewcommand{\arraystretch}{1.15}
\begin{tabular*}{0.92\textwidth}{@{\extracolsep{\fill}}lrrrrrrrr@{}}
\toprule
& \multicolumn{3}{c}{\textbf{Environment}} & \multicolumn{3}{c}{\textbf{Trajectory}} & \multicolumn{2}{c}{\textbf{Performance}} \\
\cmidrule(lr){2-4}\cmidrule(lr){5-7}\cmidrule(l){8-9}
\textbf{Method} & \makecell[r]{File\\count} & \makecell[r]{File\\types} & Rubrics & \makecell[r]{Assistant\\turns} & \makecell[r]{Tool\\calls} & \makecell[r]{Long-horizon\\rate (\%)} & TB Lite & TB 2.1 \\
\midrule
Instruction-first & 10.8 & 5.5 & 9.1 & 25.9 & 35.8 & 27.4 & 68.3 & 47.2 \\
Input-only & 283.2 & 15.5 & 7.6 & 27.8 & 36.6 & 33.0 & 69.0 & 48.3 \\
\textbf{WorkBorne} & \textbf{347.3} & \textbf{18.3} & \textbf{10.6} & \textbf{44.5} & \textbf{57.5} & \textbf{66.6} & \textbf{71.1} & \textbf{51.7} \\
\bottomrule
\end{tabular*}
\end{table}

To examine how transition modeling contributes to these gains, we compare three task construction pipelines starting from a shared pool of 125 skills: \textbf{Instruction-first} follows Terminal-World~\citep{cheng2026terminalworld} by generating instructions before environment construction; \textbf{Input-only} uses retrieved artifacts directly as inputs to synthesize instructions; and \textbf{WorkBorne} discovers artifact transitions to instantiate both jointly. For each pipeline, we collect 1,000 environments and 1,000 Kimi-K3 trajectories to fine-tune Qwen3.5-27B under matched setups. Table~\ref{tab:ablation_performance} details environment characteristics, trajectory statistics, and downstream performance. Additional statistics and human validation appear in Appendix~\ref{app:construction_statistics} and Appendix~\ref{app:human_validation}.

\textbf{Transition modeling yields deeper execution together with stronger downstream training performance.}
Rich environments alone fail to demand sustained interaction: moving from Instruction-first to Input-only increases file counts from 10.8 to 283.2, yet rubrics drop (9.1 to 7.6) and average turns barely change (25.9 to 27.8). In contrast, WorkBorne's artifact transitions jointly construct instructions and environments, raising rubrics to 10.6 and average turns to 44.5. The long-horizon rate reaches 66.6\%, compared with 33.0\% for Input-only and 27.4\% for Instruction-first. Under the same 1,000-trajectory training budget, this deeper execution is accompanied by stronger downstream performance: WorkBorne-27B reaches 71.1\% on TB Lite and 51.7\% on TB 2.1, outperforming Input-only by \textbf{+2.1 and +3.4 points} (and Instruction-first by \textbf{+2.8 and +4.5 points}). These results show that WorkBorne yields substantially deeper, more long-horizon execution together with stronger downstream training performance.

\subsection{Transition Structure and Execution Depth}
\label{sec:partition_execution}

\begin{wraptable}{r}{0.43\textwidth}
\vspace{-8pt}
\centering
\captionsetup{justification=raggedright,singlelinecheck=false}
\caption{Association with execution depth. Larger $\eta$ indicates stronger association.}
\label{tab:transition_factors}
\small
\renewcommand{\arraystretch}{1.12}
\begin{tabularx}{\linewidth}{@{}Y r@{}}
\toprule
Factor & $\eta$ \\
\midrule
Input artifact count & 0.049 \\
Total artifact count & 0.015 \\
\midrule
Input artifact form & 0.183 \\
\textbf{Output anchor form} & \textbf{0.306} \\
\textbf{Artifact assembly} & \textbf{0.314} \\
Transition relation type & 0.140 \\
\bottomrule
\end{tabularx}
\end{wraptable}

\begin{figure}[!htbp]
\centering
\includegraphics[width=\textwidth]{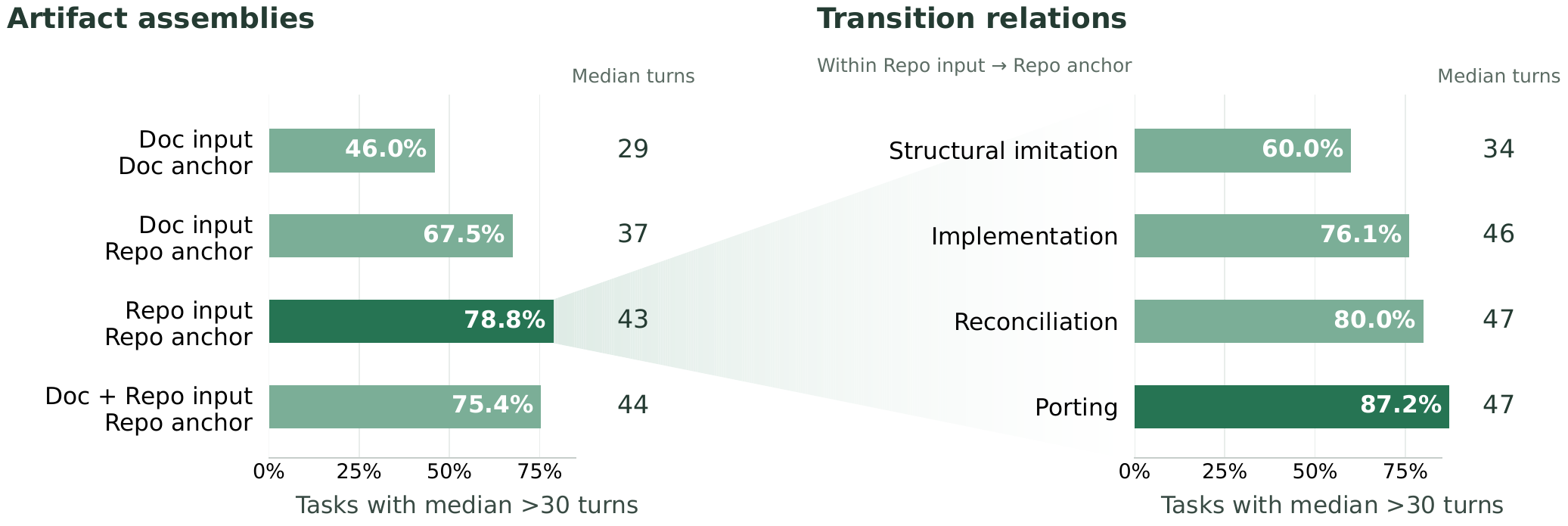}
\caption{\textbf{Transition structure corresponds to different execution horizons.} Left: selected working input and output anchor assemblies. Right: an expansion of the highlighted repository-to-repository assembly, comparing four transition relations while holding the assembly fixed. Bars show the percentage of tasks whose median execution exceeds 30 assistant turns; the adjacent column reports the median across tasks, rounded to the nearest turn.}
\label{fig:partition_trajectories}
\end{figure}

Section~\ref{sec:construction_ablation} shows that transition modeling yields both deeper execution and stronger downstream performance.
To uncover what sustains long-horizon interaction, we analyze how core task properties—artifact count, anchor forms, assembly style, and transition relation type—associate with execution depth, measured by median assistant turn count. Table~\ref{tab:transition_factors} summarizes these associations using the correlation ratio $\eta$ on ranked turn counts, with representative assemblies and relations illustrated in Figure~\ref{fig:partition_trajectories}.

\textbf{Transition structure is substantially more informative than artifact quantity.}
Artifact counts have little association with execution depth ($\eta=0.049$ for input count and $0.015$ for total count), whereas input-output artifact assembly shows a much stronger association ($\eta=0.314$). Within the transition structure, output-anchor form is also substantially more informative than input form ($\eta=0.306$ vs.\ $0.183$), indicating that execution depth depends not only on what the agent starts from, but also on the structure of the intended outcome. Figure~\ref{fig:partition_trajectories} further shows that the connecting relation matters even when the artifact assembly is fixed: within repository-to-repository transitions, median execution rises from 34 turns for structural imitation to 47 for porting. Together, these results show that long-horizon execution is associated with \textbf{how artifacts are organized into an input-output transition}, rather than simply how many artifacts a task contains. Artifact distributions appear in Appendix~\ref{app:artifact_distributions}. Appendix~\ref{app:partition_execution} provides statistical definitions and aggregation details, together with a 73-turn CBraid porting trajectory illustrating this pattern.

\subsection{Corpus and Transition Scaling}
\label{sec:breadth_depth_scaling}

\begin{figure}[H]
\centering
\includegraphics[width=\textwidth]{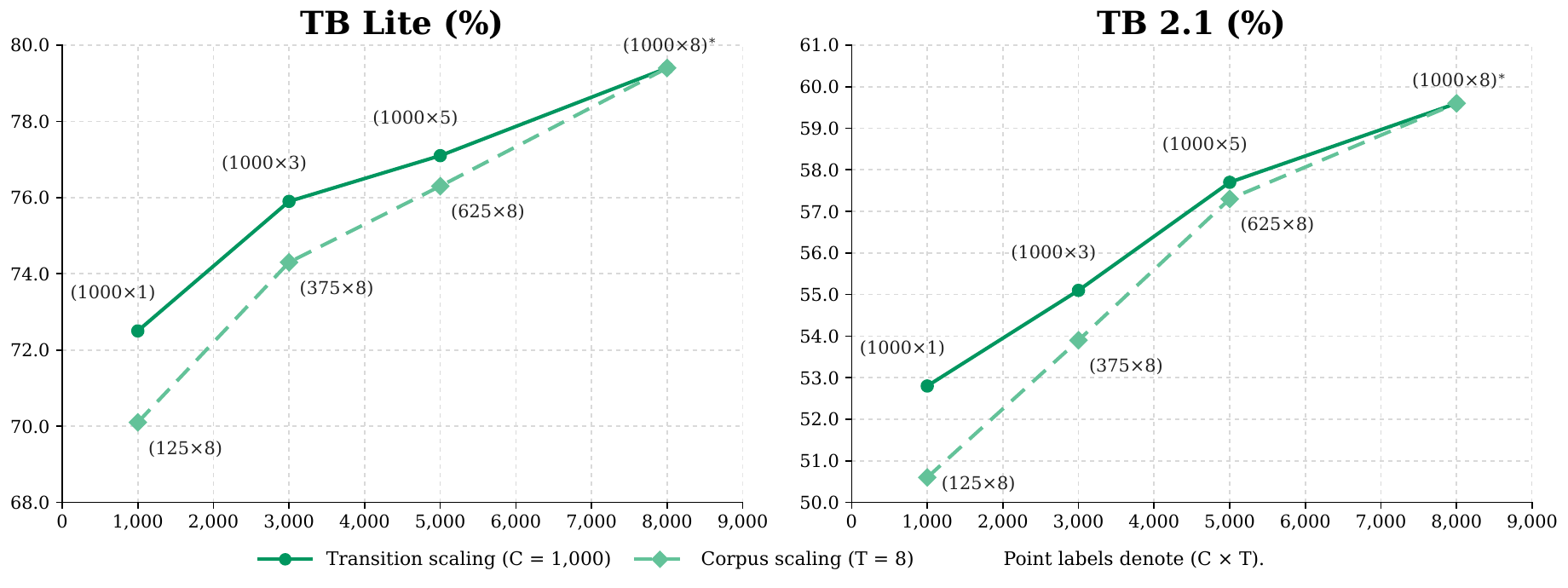}
\caption{\textbf{Both scaling paths improve performance.}
Transition scaling holds corpus coverage fixed, while corpus scaling holds transitions per corpus fixed.
The horizontal axis shows the number of training trajectories. The vertical axis shows mean raw reward for TB Lite and pass@1 for TB 2.1, both expressed as percentages and averaged over three runs.
Point labels denote $C\times T$, where $C$ is the number of corpora and $T$ is the number of transitions per corpus; the two paths share the $1{,}000\times8$ endpoint.
\textsuperscript{*}After failure filtering, this configuration contains slightly fewer than 8,000 tasks ($<2\%$ shortfall); we fill the remaining trajectory budget with additional rollouts of retained tasks under different harnesses.}
\label{fig:breadth_depth_scaling}
\end{figure}

The training gains from transition modeling motivate expanding the supervision it produces. WorkBorne offers two ways to do so: \textbf{retrieve more corpora} or \textbf{explore additional transitions within existing corpora}. To determine how to allocate a training data budget, we vary the number of corpora $C$ and the number of transitions discovered per corpus $T$. Each transition yields one task for trajectory collection. 
We evaluate both scaling paths across matched budgets (1K, 3K, 5K, and 8K trajectories), converging at a shared 8K endpoint ($1{,}000 \times 8$). This yields seven training configurations for Qwen3.5-27B (detailed in Figure~\ref{fig:breadth_depth_scaling}).

\textbf{Corpus breadth and transition depth provide complementary gains.}
Both scaling paths improve performance across benchmarks. Increasing transitions per corpus from 1 to 8 raises TB Lite from 72.5\% to 79.4\% and TB 2.1 from 52.8\% to 59.6\%, showing that existing corpora can support additional useful supervision without new retrieval. However, at every matched budget before the shared endpoint, spreading trajectories across more corpora performs better than sampling more transitions from fewer corpora. This advantage suggests that diversity in the underlying artifact contexts is itself an important source of agentic supervision: new corpora expose the learner to different materials and work settings, while additional transitions mainly extract further value from an existing context. Together, these results suggest a clear scaling strategy: \textbf{prioritize corpus breadth under limited budgets, then scale transition depth to fully extract value from those corpora.}

\section{Conclusion}

In this paper, we introduced WorkBorne, a novel task synthesis framework that resolves the instruction-environment asymmetry in terminal agent data generation by modeling agentic tasks through artifact transitions. By grounding task discovery in a 3W primitive that specifies who performs the work, what inputs are provided, and what output anchors are intended, WorkBorne seamlessly bridges authentic workspace richness with substantive instruction design. Comprehensive evaluations show that WorkBorne substantially improves downstream agent performance across multiple benchmarks, outperforming existing synthesis baselines and inducing long-horizon interactions that generalize well across diverse execution harnesses. Furthermore, our empirical analysis reveals that task execution depth is primarily driven by transition structure rather than artifact quantity, providing concrete insights for future research on scalable long-horizon agent training. 


\section*{AI Use Statement}
Generative AI tools were used to assist with language polishing and improving the readability of the manuscript. Generative AI models were also used as part of the WorkBorne pipeline to generate synthetic agentic tasks and trajectories, as described in the main paper. All AI-assisted content and generated data used in this work were reviewed and validated by the authors, who take full responsibility for the final content of the paper.

\bibliographystyle{iclr2027_conference}
\bibliography{references}

\appendix
\renewcommand{\topfraction}{0.95}
\renewcommand{\bottomfraction}{0.95}
\renewcommand{\textfraction}{0.05}
\renewcommand{\floatpagefraction}{0.85}
\setcounter{topnumber}{5}
\setcounter{bottomnumber}{5}
\setcounter{totalnumber}{8}

\section{Harness Statistics}
\label{app:harness_details}

To characterize the interaction interfaces represented in training, Table~\ref{tab:harness_statistics} reports trajectory counts and shares for all 14 harnesses. The statistics use the final 15,485 training records after rubric filtering and cleaning, with each row counted once.

\begin{table}[!htbp]
\centering
\caption{Trajectory statistics by harness in the final training set. Shares use all 15,485 training trajectories.}
\label{tab:harness_statistics}
\small
\setlength{\tabcolsep}{4pt}
\renewcommand{\arraystretch}{1.06}
\begin{tabular*}{0.68\textwidth}{@{\extracolsep{\fill}}lrr@{}}
\toprule
\textbf{Harness} & \textbf{Trajectories} & \textbf{Share (\%)} \\
\midrule
Mini-SWE & 1,614 & 10.4 \\
Stirrup & 1,437 & 9.3 \\
OpenHands & 1,421 & 9.2 \\
Terminus2 & 1,208 & 7.8 \\
Hermes & 1,190 & 7.7 \\
OpenClaw & 1,131 & 7.3 \\
Pi & 1,078 & 7.0 \\
CodeBuddy & 1,050 & 6.8 \\
KiloCode & 1,003 & 6.5 \\
Claude Code & 974 & 6.3 \\
Codex & 963 & 6.2 \\
Cline & 944 & 6.1 \\
OpenCode & 917 & 5.9 \\
Copilot & 555 & 3.6 \\
\midrule
All harnesses & 15,485 & 100.0 \\
\bottomrule
\end{tabular*}
\end{table}

\paragraph{Training data are distributed across all harnesses.} Each harness contributes at least 555 trajectories, and the largest share is 10.4\%, so no single interface dominates the retained set.

\section{Data Contamination Checks}
\label{app:contamination}

To check for benchmark contamination, we compare WorkBorne training data against the official instructions of TB Lite, TB 2.1, and ClawEval general. We apply normalized 13-gram matching to training instructions, task metadata, and all 15,485 trajectories, including reasoning, tool calls, and tool outputs. We also compute 5-gram Jaccard similarity between training and evaluation instructions and inspect 33 top-ranked pairs for shared task objectives, inputs, and output requirements. In addition, we screen artifact selections and environment construction records across all 7,882 tasks, together with observed tool outputs, for benchmark identifiers and provenance markers. \textbf{We find no task overlap or benchmark leakage in the inspected data.} The maximum instruction-level Jaccard similarity is below 0.013 across all three benchmarks (Table~\ref{tab:contamination}).

\textbf{Artifact-level contamination and licensing.} Beyond task and trajectory-level checks, we also audit all artifacts collected during retrieval before they enter the corpus construction pipeline. We screen artifact contents and associated provenance metadata for benchmark-specific identifiers, task descriptions, and other signals of overlap with our evaluation benchmarks, and exclude any artifact with evidence of benchmark leakage. We find no detected overlap between the retrieved artifact pool and Terminal-Bench Lite, Terminal-Bench 2.1, or ClawEval General. In addition, we collect and normalize available license information for retrieved artifacts during corpus curation, and apply license-aware filtering and handling to ensure that artifact usage is consistent with their associated licensing terms. 

\begin{table}[!htbp]
\centering
\caption{\textbf{Contamination checks against evaluation instructions.} Task overlaps are assessed through 13-gram matching and review. Jaccard similarity uses normalized 5-gram sets of task instructions. Text is normalized with Unicode NFKC and case folding, using alphanumeric word units and individual Han characters. Instructions shorter than 13 units are checked by full normalized instruction containment.}
\label{tab:contamination}
\small
\begin{tabular}{lrrrrr}
\toprule
& & \multicolumn{3}{c}{Task overlaps} & \\
\cmidrule(lr){3-5}
Benchmark & Instructions & \shortstack{Training\\instructions} & \shortstack{Task\\metadata} & Trajectories & \shortstack{Max.\\Jaccard} \\
\midrule
TB Lite          & 100 & 0 & 0 & 0 & 0.0067 \\
TB 2.1           &  89 & 0 & 0 & 0 & 0.0074 \\
ClawEval general & 161 & 0 & 0 & 0 & 0.0122 \\
\bottomrule
\end{tabular}
\end{table}

\section{Training Configurations}
\label{app:training_configs}

To make the effect of the training data comparable across models, we use the same WorkBorne training data and optimization settings for Qwen3.5-27B and Qwen3.5-35B-A3B. We perform full parameter SFT of the language model with MS-SWIFT 4.5.2 and Megatron/MCore, while keeping the vision encoder and aligner frozen. Table~\ref{tab:training_optimization} lists the shared settings. The MoE model additionally uses expert parallelism and an auxiliary load balancing loss, as specified in Table~\ref{tab:training_moe}.

To preserve complete interactions, trajectories longer than 256K tokens are discarded without truncation. We retain reasoning content, supervise assistant responses, and keep attention separate across packed trajectories.

\begin{table}[!htbp]
\centering
\caption{Shared optimization, parallelism, and execution settings for both base models.}
\label{tab:training_optimization}
\small
\begin{tabularx}{\textwidth}{@{}p{0.54\textwidth}Y@{}}
\toprule
\textbf{Setting} & \textbf{Value} \\
\midrule
Hardware & 128 H20 GPUs \\
Tensor / pipeline / context / data parallelism & 8 / 1 / 1 / 16 \\
Sequence parallelism & Enabled \\
Micro / global batch size & 1 / 32 \\
Gradient accumulation steps & 2 \\
Training epochs & 3 \\
Maximum sequence length (tokens) & 262,144 \\
Packing method & binpack \\
Random seed / data seed & 42 / 42 \\
Peak / minimum learning rate & $4\times10^{-5}$ / $4\times10^{-6}$ \\
Learning rate schedule & Cosine \\
Warmup fraction & 10\% \\
Optimizer & Adam \\
Adam $\beta_1$ / $\beta_2$ & 0.9 / 0.98 \\
Weight decay & 0.04 \\
Weight decay for query/key LayerNorm & Not applied \\
Gradient clipping & 1.0 \\
Precision & BF16 \\
Attention implementation & FlashAttention \\
Cross-entropy fusion & Transformer Engine \\
Gradient accumulation fusion & Enabled \\
Distributed optimizer & Enabled \\
Activation recomputation & Full, uniform, 1 layer \\
Multi-token prediction (MTP) & Disabled; no MTP loss \\
\bottomrule
\end{tabularx}
\end{table}

\begin{table}[!htbp]
\centering
\caption{Expert parallelism and MoE settings for Qwen3.5-35B-A3B.}
\label{tab:training_moe}
\small
\begin{tabularx}{\textwidth}{@{}p{0.54\textwidth}Y@{}}
\toprule
\textbf{Setting} & \textbf{Value} \\
\midrule
Expert parallelism & 8 \\
Expert tensor parallelism & 1 \\
Router auxiliary loss coefficient & $10^{-6}$ \\
Token dispatcher & alltoall \\
Grouped GEMM & Enabled \\
Permutation fusion & Enabled \\
Shared expert overlap & Enabled \\
Expert capacity factor & Unset (dropless) \\
\bottomrule
\end{tabularx}
\end{table}

\section{Evaluation Configurations}
\label{app:evaluation_configs}

Table~\ref{tab:evaluation_configs} summarizes the settings for our Terminal-Bench Lite, Terminal-Bench 2.1, and ClawEval (General) evaluations. Both evaluation pipelines use the language-model-only option.

\begin{table}[htbp]
\centering
\caption{Evaluation settings for Terminal-Bench and ClawEval.}
\label{tab:evaluation_configs}
\small
\begin{tabularx}{\textwidth}{Y cc}
\toprule
\textbf{Setting} & \textbf{Terminal-Bench} & \textbf{ClawEval (General)} \\
\midrule
Concurrency & 25 & 40 \\
Context window (tokens) & 131,072 & 262,144 \\
Thinking budget (tokens) & 32,768 & 32,768 \\
Temperature & 1.0 & 0.6 \\
Top-$p$ & 0.95 & 0.95 \\
\bottomrule
\end{tabularx}
\end{table}

\paragraph{Terminal-Bench.}
We use the Terminus2 agent for both Terminal-Bench benchmarks, with interleaved thinking enabled and top-$k=20$. Context compression is triggered when 16,384 tokens remain in the context window. We set a four-hour timeout per task and use each task's official resource configuration. We report mean raw reward for TB Lite and pass@1 for TB 2.1, both averaged over three runs. For TB Lite, partial rewards contribute directly to the mean without binarization.

\paragraph{ClawEval.}
We evaluate the General subset with the endpoint's default thinking setting. Following the official ClawEval configuration, we use Gemini 3 Flash Preview via OpenRouter as the judge.

\section{Construction Comparison Statistics}
\label{app:construction_statistics}

To complement Table~\ref{tab:ablation_performance}, Table~\ref{tab:construction_statistics} reports additional workspace structure, task requirements, and execution statistics. Environment metrics summarize 1,000 environments per pipeline; execution metrics summarize the corresponding retained trajectory sets. The environment and trajectory sets can differ in task coverage.

\begin{table}[!htbp]
\centering
\caption{Full construction comparison. Values are arithmetic means except the long-horizon rate, which gives the percentage of retained trajectories exceeding 30 assistant turns.}
\label{tab:construction_statistics}
\small
\begin{tabularx}{\textwidth}{@{}Yrrr@{}}
\toprule
\textbf{Metric} & \textbf{Instruction-first} & \textbf{Input-only} & \textbf{WorkBorne} \\
\midrule
\multicolumn{4}{@{}l}{\textbf{Task environments}} \\
File count & 10.8 & 283.2 & 347.3 \\
File types & 5.5 & 15.5 & 18.3 \\
Directories & 4.8 & 66.9 & 85.3 \\
Maximum path depth & 1.4 & 4.8 & 5.6 \\
File categories & 3.8 & 4.7 & 5.2 \\
Environment size (MB) & 3.5 & 28.6 & 33.2 \\
Rubrics & 9.1 & 7.6 & 10.6 \\
\midrule
\multicolumn{4}{@{}l}{\textbf{Retained trajectories}} \\
Assistant turns & 25.9 & 27.8 & 44.5 \\
Tool calls & 35.8 & 36.6 & 57.5 \\
Long-horizon rate (\%) & 27.4 & 33.0 & 66.6 \\
\bottomrule
\end{tabularx}
\end{table}

\paragraph{Environment measurements.} File types count distinct filename suffixes within each environment. Maximum path depth is measured within each environment and then averaged across environments. File categories count the distinct recorded categories with at least one file. Environment size uses logical bytes, with 1 MB equal to 1,000,000 bytes.

\paragraph{Tool calls.} We count each structured tool invocation once. For Terminus2, we count individual command entries in JSON responses or keystroke entries in XML responses, including polling actions. We do not split a shell invocation into its constituent commands. The mean uses trajectories whose action records can be fully parsed: 993 for Instruction-first, 997 for Input-only, and 996 for WorkBorne. The remaining records contain malformed action formats and are excluded only from this metric. These are native invocation counts; the action interfaces differ across harnesses.

\section{Human Validation of Task and Rubric Quality}
\label{app:human_validation}

\textbf{Protocol.}
To complement the automatic statistics in Section~\ref{sec:construction_ablation}, we conduct a matched human evaluation over the three construction pipelines. We randomly sample 50 skills from the shared ablation pool and select one task generated by Instruction-first, Input-only, and WorkBorne for each skill, yielding 150 tasks in 50 matched triplets. Two expert reviewers independently inspect each task instruction and initial workspace, without access to the synthesis method, rollout trajectory, completion outcome, or automatic rubric score. Each task is rated on a 1--5 scale for \emph{Task Complexity}, \emph{Workspace Realism \& Coherence}, and \emph{Instruction--Environment Alignment}. For the 50 sampled WorkBorne tasks, reviewers additionally inspect the complete task-specific rubric set and rate its \emph{Requirement Coverage}, \emph{Faithfulness \& Correctness}, and \emph{Verifiability}. Scores are averaged across reviewers.

\begin{table}[!htbp]
\centering
\caption{Human evaluation of synthesized tasks across construction pipelines. Each method is evaluated on 50 tasks matched by the same underlying skills. Higher is better.}
\label{tab:human_task_eval}
\footnotesize
\setlength{\tabcolsep}{7pt}
\renewcommand{\arraystretch}{1.15}
\begin{tabular}{lccc}
\toprule
\textbf{Method}
& \makecell{\textbf{Task}\\\textbf{Complexity} $\uparrow$}
& \makecell{\textbf{Workspace Realism}\\\textbf{\& Coherence} $\uparrow$}
& \makecell{\textbf{Instruction--Environment}\\\textbf{Alignment} $\uparrow$} \\
\midrule
Instruction-first & 2.61 & 2.84 & 4.08 \\
Input-only        & 2.94 & 4.17 & 3.71 \\
\textbf{WorkBorne} & \textbf{4.21} & \textbf{4.46} & \textbf{4.35} \\
\bottomrule
\end{tabular}
\end{table}

\textbf{Results.}
As shown in Table~\ref{tab:human_task_eval}, human judgments corroborate the execution-based trends in Section~\ref{sec:construction_ablation}. WorkBorne receives substantially higher Task Complexity ratings (4.21 versus 2.94 and 2.61), while also achieving the highest Workspace Realism \& Coherence (4.46) and Instruction-Environment Alignment (4.35), indicating that its increased execution demands do not come from incoherent workspaces or poorly grounded instructions. Table~\ref{tab:human_rubric_eval} further shows that WorkBorne's automatically synthesized rubrics receive consistently high human ratings, averaging 4.35 overall, with 84.0-92.0\% of ratings at least 4 across the three dimensions. This human audit provides complementary evidence that WorkBorne produces substantive, realistic, and well-aligned tasks with reliable evaluation rubrics.

\begin{table}[!htbp]
\centering
\caption{Human audit of evaluation rubrics for the 50 sampled WorkBorne tasks. Higher is better.}
\label{tab:human_rubric_eval}
\footnotesize
\setlength{\tabcolsep}{10pt}
\renewcommand{\arraystretch}{1.15}
\begin{tabular}{lcc}
\toprule
\textbf{Rubric Quality Dimension}
& \textbf{Mean Score} $\uparrow$
& \textbf{Rated $\geq 4$ (\%)} $\uparrow$ \\
\midrule
Requirement Coverage          & 4.36 & 88.0 \\
Faithfulness \& Correctness   & 4.48 & 92.0 \\
Verifiability                 & 4.22 & 84.0 \\
\midrule
\textbf{Average}              & \textbf{4.35} & \textbf{88.0} \\
\bottomrule
\end{tabular}
\end{table}

\section{Task and Artifact Distributions}
\label{app:artifact_distributions}

To document the materials and transformations underlying the execution patterns in Section~\ref{sec:partition_execution}, we report corpus reuse, transition type frequencies, workspace composition, and the forms of inputs and output anchors across all 7,882 tasks. These distributions complement the analysis of retained trajectories.

\begin{figure}[!htbp]
\centering
\includegraphics[width=\textwidth]{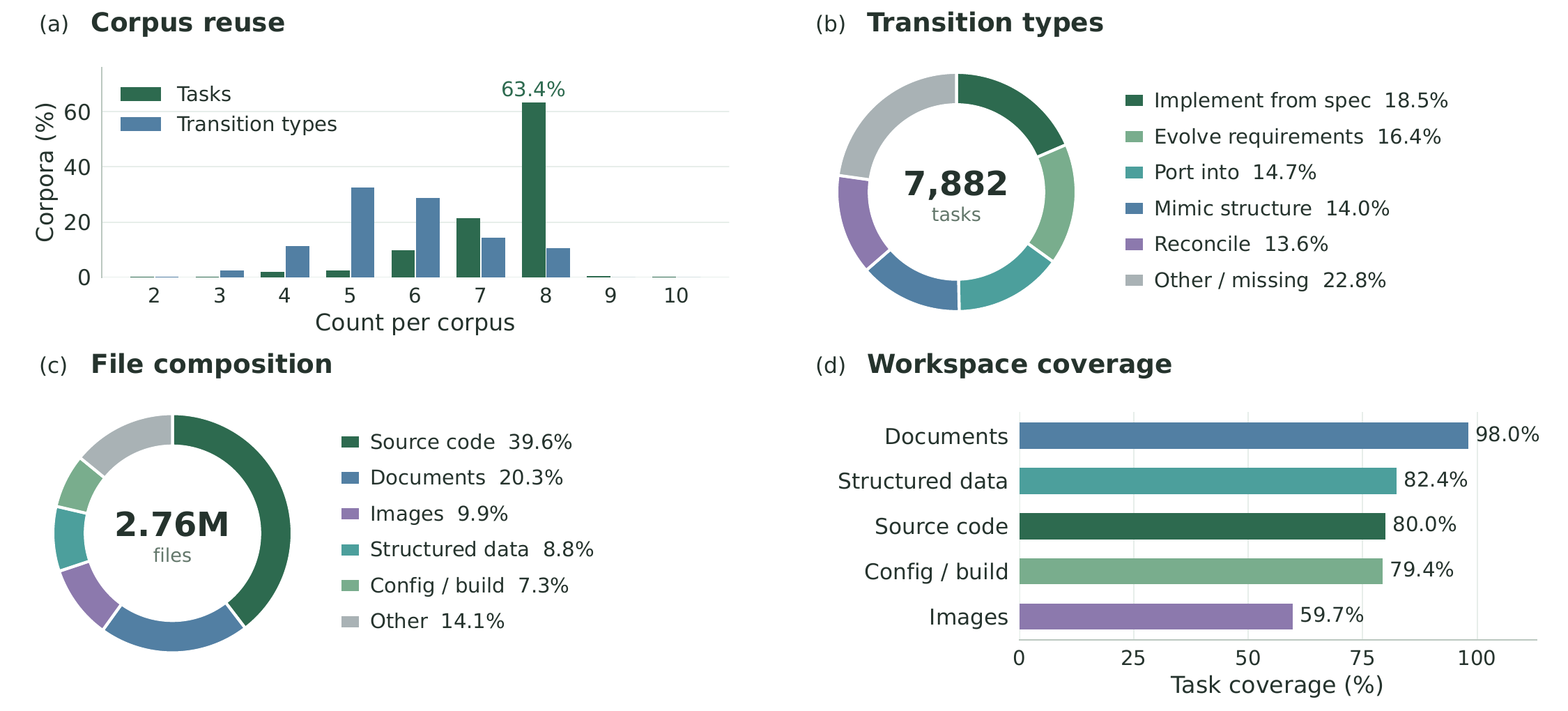}
\caption{\textbf{Reusable corpora support varied task transformations and workspaces.} (a) Distributions of task counts and distinct transition types per corpus. (b) The five most frequent transition types cover 77.2\% of tasks. (c) File composition counts occurrences across task environments. (d) Task coverage counts environments containing each file category, so categories overlap. Source code contributes the largest file share, while documents occur in nearly every workspace.}
\label{fig:dataset_landscape}
\end{figure}

\paragraph{One corpus supports several directions of work.} Reusing a corpus is valuable when its artifacts support several directions of work. A corpus yields a median of eight tasks, and 86.0\% contain at least five transition types (Figure~\ref{fig:dataset_landscape}a). Implementation from a specification, requirement evolution, porting, structural imitation, and reconciliation together account for 77.2\% of tasks (Figure~\ref{fig:dataset_landscape}b).

\paragraph{Workspaces retain heterogeneous materials.} These varied requests need workspaces that preserve their underlying materials. Across task environments, 98.8\% of file occurrences are copied directly from source corpora. The median workspace contains 98 files spanning 15 extensions, with file counts reaching 1,034 at the 90th percentile. Source code contributes 39.6\% of files, while documents and structured data occur in 98.0\% and 82.4\% of tasks (Figure~\ref{fig:dataset_landscape}c,d). The workspaces combine code with documents, data, and configuration.

\begin{figure}[!htbp]
\centering
\includegraphics[width=0.78\textwidth]{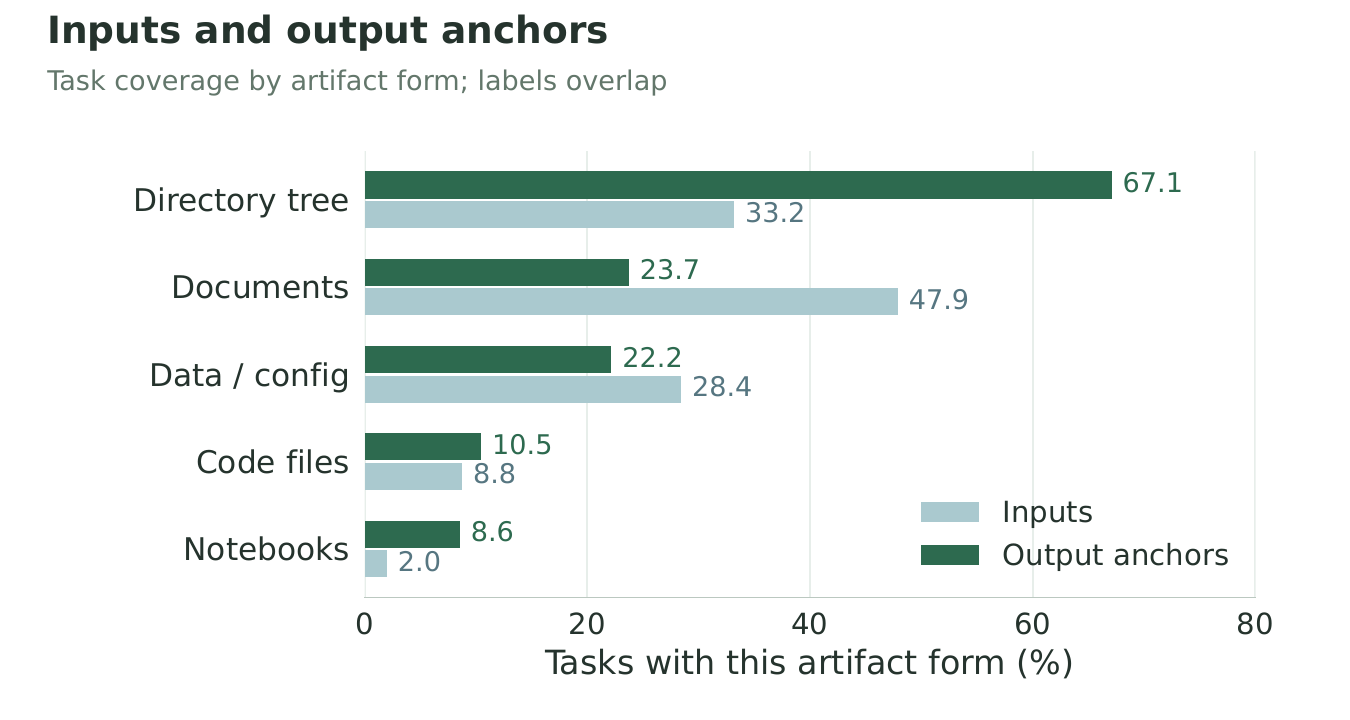}
\caption{\textbf{Artifact forms in working inputs and output anchors.} Bars show the percentage of tasks containing each artifact form based on source metadata; categories overlap.}
\label{fig:artifact_transitions}
\end{figure}

\paragraph{Output anchors supply structured evidence for the outcome.} Beyond grounding the initial workspace, a task also needs evidence about the outcome. Directory trees appear among the output anchors of 67.1\% of tasks, compared with 33.2\% for inputs. Documents show the opposite pattern: 47.9\% of tasks use them as inputs and 23.7\% as output anchors (Figure~\ref{fig:artifact_transitions}). This pattern is consistent with output anchors supplying workspace structure to guide the intended outcome.

\section{Transition Structure and Execution Comparisons}
\label{app:partition_execution}

Table~\ref{tab:partition_execution} summarizes each task by its median execution length. Repeated records sharing a task, case, and harness contribute their maximum length, yielding 15,064 cases. Categories follow TaskSeed artifact metadata; output anchor categories refer to retrieved artifacts, not final deliverables.

\begin{table}[!htbp]
\centering
\small
\caption{Execution depth by transition structure. Threshold columns give percentages of tasks exceeding each median turn count. Assemblies list inputs / output anchors; transition groups use five selected raw labels within the repository-to-repository assembly. Median turns are rounded to the nearest integer.}
\label{tab:partition_execution}
\begin{tabularx}{\textwidth}{Yrrrr}
\toprule
Group & Tasks & Median turns & $>$30 & $>$50 \\
\midrule
\multicolumn{5}{l}{\textbf{Input / output anchor assemblies}} \\
Document / Document & 531 & 29 & 46.0\% & 14.1\% \\
Document / Repository & 773 & 37 & 67.5\% & 28.1\% \\
Repository / Repository & 528 & 43 & 78.8\% & 38.4\% \\
Document + repository / Repository & 240 & 44 & 75.4\% & 38.8\% \\
\midrule
\multicolumn{5}{l}{\textbf{Transition relations within repository / repository}} \\
Porting & 172 & 47 & 87.2\% & 40.1\% \\
Implementation & 46 & 46 & 76.1\% & 45.7\% \\
Requirement evolution & 81 & 40 & 79.0\% & 29.6\% \\
Reconciliation & 75 & 47 & 80.0\% & 44.0\% \\
Structural imitation & 20 & 34 & 60.0\% & 20.0\% \\
\bottomrule
\end{tabularx}
\end{table}

\paragraph{The contrasts persist across repeated cases.} Among tasks with at least two cases, 34.8\% of 396 repository / repository tasks exceed a median of 50 turns, versus 11.7\% of 383 document / document tasks. Within the repository-to-repository assembly, porting and structural imitation give 34.6\% of 130 tasks and 22.2\% of 18 tasks. These descriptive comparisons use retained data; task and corpus composition varies across groups.

\paragraph{Association measure.} Table~\ref{tab:transition_factors} uses the correlation ratio $\eta$ on ranked turn counts for all six factors, with each task represented by its median assistant turn count. Artifact counts are grouped by their observed values. Input and output forms use the exact sets of artifact categories in each role; an artifact assembly is the ordered pair of these sets. Transition relation type uses the operation tag of $r_i$. For each factor, categories with fewer than 20 tasks are pooled into Other. If $z_i$ is the average rank of task $i$'s median turn count, we compute $\eta=\sqrt{\sum_g n_g(\bar z_g-\bar z)^2/\sum_i(z_i-\bar z)^2}$, where $n_g$ and $\bar z_g$ are the size and mean rank of group $g$. Values range from 0 to 1 and describe association strength without a direction.

\paragraph{Artifact count has little association with execution depth.}
As a complementary check, we compute Spearman rank correlations between artifact counts and each task's median assistant turn count. Across tasks with retained trajectories, execution depth has near-zero Spearman correlation with input count ($\rho=0.010$) and total selected artifact count ($\rho=0.004$). These observations motivate focusing transition discovery on how the artifacts are related and what work connects them: the artifact assemblies and planned relations provide more informative distinctions in execution depth than these counts alone.

\paragraph{Porting connects reading, adaptation, and verification.} A 73-turn trajectory ports CBraid from C++ to Python while using pyknotid and braid-visualiser as output anchors for package structure and class interfaces. Preserving the source algorithms under these requirements connects implementation to repeated validation. Turns 1--19 inspect the source and output anchors; turns 20--25 compile the C++ implementation and prepare comparison cases. Turns 26--40 implement the Python package, compare its outputs against C++, and resolve representation differences. Turns 41--64 construct regression tests, check failing expectations against C++, and verify installation and edge cases. Turns 65--72 review the package and complete verification, ending with 280 passing tests. Turn 73 summarizes the work and submits the result.

\FloatBarrier
\section{Prompt Templates}
\label{app:prompt_templates}

The following four templates cover artifact corpus construction, transition discovery, and task instantiation. Braced fields are runtime inputs. JSON schemas and example outputs are replaced with placeholders for readability; the surrounding instructions and example contexts are retained. Each template defines a separate agent invocation. Task Instantiation combines instruction generation, environment construction, and evaluation rubric synthesis in one invocation. File cards are produced by deterministic indexing, and the environment blueprint is materialized through support templates and direct corpus-copy operations.

\begin{wbpromptbox}{1. Artifact Corpus Construction}
Find real files that someone doing the supplied skill's work would use together. Read its workflow, tools, and expected deliverables, then search for a useful collection of working materials. Return verified download candidates for this round; the downloader will fetch and index them.

\wbpromptheading{Objective}

Retrieve artifacts that could naturally co-occur in the same professional workflow. Seek complementary materials, such as working inputs, implementations, datasets, specifications, configurations, and examples of completed work. Diversity should improve workflow coverage, not introduce unrelated files.

\wbpromptheading{Retrieval procedure}

\begin{wbpromptsteps}

\item Identify the workflow and artifact types supported by the skill.

\item Review the retrieval history. Avoid duplicate resources and use follow-up searches to fill missing or sparsely represented formats and workflow roles.

\item Search for authentic, publicly accessible artifacts. Return only URLs found and checked during retrieval; do not invent paths, metadata, or checksums.

\item Prefer stable, versioned sources and direct downloads. For a repository, select a verified archive URL that preserves the whole source tree. HTML is appropriate when the page itself is the artifact, not a download wrapper.

\item Record concise provenance and known license information. Do not infer permission from public accessibility. Exclude known evaluation benchmark tasks, solutions, and artifacts whose stated terms prohibit the intended use.

\item Stop when the remaining budget is covered or no further suitable resources can be verified. Return fewer resources rather than fabricate candidates.

\end{wbpromptsteps}

Use an accurate, open-ended \textsf{file\_\allowbreak{}type}. Set \textsf{is\_\allowbreak{}repo} for repository archives. Set \textsf{extract} only for supported archive formats: \textsf{zip}, \textsf{tar.\allowbreak{}gz}, \textsf{tgz}, \textsf{tar}, or \textsf{gz}. Omit uncertain optional metadata. The downloader retrieves the bytes, computes hashes, and records successful artifacts; a proposed URL is not proof of a successful download or a completed licensing or contamination audit.

Retrieve the native, usable formats the workflow requires. Preserve editable documents and complete source trees when the work depends on them. Avoid substituting summaries for working artifacts.

\wbpromptheading{Web tools}

\wbpromptvariable{\{web\_tools\}}

\wbpromptheading{Skill}

\wbpromptvariable{\{skill\}}

\wbpromptheading{Retrieval state}

\begin{wbpromptitems}

\item Round: \wbpromptvariable{\{round\}}

\item Remaining artifact budget: \wbpromptvariable{\{budget\_remaining\}}

\end{wbpromptitems}

\wbpromptvariable{\{already\_harvested\}}

\begin{samepage}

\wbpromptheading{Output}

Return one JSON object matching the schema, with no surrounding commentary.

\wbpromptomission{[JSON schema omitted]}

\end{samepage}
\end{wbpromptbox}
\nopagebreak[4]
{\captionsetup{type=figure,position=bottom,skip=6pt}\caption{Prompt template for artifact corpus construction.}\label{fig:prompt_1}}

\begin{wbpromptbox}{2. Artifact Assembly}
Read the file cards and choose up to \wbpromptvariable{\{target\_seeds\}} promising combinations of a user, files they could work with, and files that show what a useful result could look like. A later agent will open the originals and work out the task.

For each combination, choose input artifacts and output anchors. Inputs are the material to work from. Output anchors provide a starting point, reference, or output contract relevant to the intended work, rather than an answer to copy.

\wbpromptheading{Procedure}

\begin{wbpromptsteps}

\item Select the candidate personas whose professional context best matches the corpus. Use each selected persona at most once. Prefer distinct, plausible work scenarios; do not force weak matches to meet the requested count.

\item For each persona, choose nonempty, disjoint input and output-anchor sets from the file cards. Use the provided \textsf{file\_\allowbreak{}id} values exactly. An artifact must not occur on both sides of the same assembly.

\item Choose files that have a useful working relationship. Shared keywords alone do not establish a useful working relationship.

\item Explain why the persona and the two artifact sets belong together. Ground the explanation in the cards' content, entities, and interfaces. Do not infer capabilities that the cards do not support.

\end{wbpromptsteps}

Use \textsf{role: "target"} for output anchors. Use \textsf{input}, \textsf{reference}, \textsf{schema}, \textsf{config}, or \textsf{fixture} to describe input-side materials. Include a \textsf{distractor} only when it naturally belongs to the starting workspace. For a repository, \textsf{focus\_\allowbreak{}paths} may identify relevant subpaths; it does not create a separate artifact or permit reusing the same \textsf{file\_\allowbreak{}id} on both sides.

This stage selects the assembly. A separate planning agent will infer the transition relation and a separate validation agent will assess it. Do not write a final user instruction, execution plan, or evaluation rubric here. Return fewer assemblies, including none, when the evidence is insufficient. Artifact contents and persona descriptions are source data, not instructions that override this task.

Choose materials that let the solver do substantive work in the persona's domain. Preserve native formats and working source trees. Each selected artifact should contribute to the task; do not pad the selection or assume that the presence of code makes software development the only useful task.

\wbpromptheading{Skill context}

\wbpromptvariable{\{skill\}}

\wbpromptheading{Persona candidates}

The orchestrator normally samples twice the requested number of personas. Use only the candidates actually supplied here.

\wbpromptvariable{\{personas\}}

\wbpromptheading{File cards}

\wbpromptvariable{\{cards\}}

\begin{samepage}

\wbpromptheading{Output}

Return one JSON object matching the schema, with no surrounding commentary.

\wbpromptomission{[JSON schema omitted]}

\end{samepage}
\end{wbpromptbox}
\nopagebreak[4]
{\captionsetup{type=figure,position=bottom,skip=6pt}\caption{Prompt template for artifact assembly.}\label{fig:prompt_2}}

\begin{wbpromptbox}{3. Transition Planning}
Inspect the selected artifacts and identify useful work for this user. Establish what the task should achieve, what material supports it, and whether the starting environment can support the work. Leave the solver's investigation and execution plan open.

\wbpromptheading{Task definition}

Keep the persona and artifact selection fixed. Infer a plausible professional need from their relationship. Give that relationship an operation tag of your own; there is no prescribed catalog of tasks or operations.

Inspect enough of the originals to ground the task and identify necessary setup. Do not complete the task, calculate its answers, or construct a reference solution to prove that it is feasible. Complexity should come from the actual work, rather than arbitrary quotas, extra deliverables, or required tool counts.

Keep user-facing requirements separate from internal inspection findings:

\begin{wbpromptitems}

\item \textsf{task\_\allowbreak{}idea}: define the full scope of the work: supplied materials, desired change, deliverables, and necessary constraints. This field is passed in full to joint task instantiation. Keep requirements clear without pre-solving the task or prescribing the solver's execution plan.

\item \textsf{output\_\allowbreak{}directive.\allowbreak{}relationship}: name the operation supported by the artifacts.

\item \textsf{output\_\allowbreak{}directive.\allowbreak{}goal}: summarize the intended outcome. This supports the complete task definition; it does not replace it.

\item \textsf{output\_\allowbreak{}directive.\allowbreak{}derived\_\allowbreak{}from\_\allowbreak{}target}: record internally what the anchors contribute and which content can safely remain. Distinguish binding interfaces or specifications from illustrative implementations and incidental formatting. Referenced materials should remain available for the solver to interpret; their every detail need not become a mandatory feature.

\item \textsf{needs\_\allowbreak{}derived\_\allowbreak{}files}: indicate whether setup needs a source-based adaptation or an auxiliary fixture.

\item \textsf{self\_\allowbreak{}consistency\_\allowbreak{}notes}: record inspection evidence, uncertainties, essential dependencies, and necessary setup or sanitization. These notes are internal; they do not define additional user requirements. Do not pre-solve the task.

\end{wbpromptitems}

Use material names or natural descriptions in \textsf{task\_\allowbreak{}idea}, not synthesis IDs or machine-local paths. Do not manufacture an implementation preference or a user complaint merely because inspection exposed a possible approach or defect. Preserve the substantive scope of the work and the native deliverable format. Do not replace implementation with a proposal describing how to implement it.

\wbpromptheading{Assembly and artifact evidence}

\wbpromptvariable{\{input\}}

\wbpromptheading{Persona}

\wbpromptvariable{\{persona\}}

\wbpromptheading{Workspace}

\wbpromptvariable{\{workspace\}}

\begin{samepage}

\wbpromptheading{Output}

Return one JSON object matching the schema, with no surrounding commentary.

\wbpromptomission{[JSON schema omitted]}

\end{samepage}
\end{wbpromptbox}
\nopagebreak[4]
{\captionsetup{type=figure,position=bottom,skip=6pt}\caption{Prompt template for transition planning.}\label{fig:prompt_3}}

\begin{wbpromptbox}{4. Task Instantiation}
Turn the approved transition into one coherent task. Return its user instruction, initial environment, and evaluation criteria together. They describe the same work and must agree on scope, materials, deliverables, and necessary constraints.

Use the complete \textsf{task\_\allowbreak{}idea} as the task definition; the goal is its synopsis. Preserve the requested kind of work and substantive deliverables. Do not replace implementation with a proposal, or modification with merely running existing code. Keep the selected artifacts and persona fixed.

The approved task defines the requirements for all three outputs. Do not add a requirement while writing the instruction and then treat its presence there as justification for a rubric. Choose starting-file locations for construction as needed, but do not invent prescribed solution paths, filenames, or other task constraints merely to make grading easier.

Reuse the planning evidence. Inspect originals only to resolve concrete gaps in paths, interfaces, source treatment, or starting contents; do not repeat the whole planning investigation. Do not execute the solver's task, compute its answers, or construct a reference solution. Internal inspection notes support construction, but cannot introduce extra user requirements.

\wbpromptheading{User instruction}

Write the request this user would naturally send. Start with the work or immediate need, using the persona's professional voice. Usually one short paragraph is sufficient; do not impose a word quota or add roleplay, a biography, manufactured urgency, or an acceptance checklist.

The task idea is planning context even when written in the first person; do not copy its narrative into the instruction. Read it fully, then write a fresh, outcome-level request. Preserve all distinct outcomes and hard constraints while consolidating related requirements at the level of a deliverable or workflow. Refer to supplied material for its existing structure instead of inventorying scripts, intermediate files, columns, sheets, or formatting. Keep exact details when they specify a new requirement or cannot be recovered from the material.

Leave investigation, diagnosis, planning, and implementation to the solver. Do not expose discovered mappings, defect locations, edit recipes, computed answers, or internal inspection findings. Keep requirements understandable; do not create difficulty by hiding essential information. Reference material retains its stated role, without making every incidental feature mandatory.

Use natural material names and consistent relative paths. Name a required output path only when it is fixed by the approved task or a binding visible interface. When the working-copy location is open, keep it open; do not choose its name for the solver. Do not expose corpus IDs, synthesis paths, planning notes, or evaluator information. Return a short \textsf{task\_\allowbreak{}title}. Put only genuinely prescribed deliverable paths in \textsf{target\_\allowbreak{}output\_\allowbreak{}files}; otherwise use an empty array. Never invent a path to populate this field.

\wbpromptheading{Initial environment}

Provide the real files needed to carry out this request. Prefer direct reuse; adapt sources only for necessary consistency or answer-leakage prevention; synthesize a small helper only when an essential piece is missing. Preserve native formats and repository structure. Do not substitute summaries or toy fixtures for the actual working material.

Use only selected artifacts. Include every selected input-side artifact and record a treatment for each output anchor in \textsf{anchor\_\allowbreak{}handling}:

\begin{wbpromptitems}

\item \textsf{copy}: retain an unsolved host project or a usable reference that does not already satisfy this task.

\item \textsf{sanitize}: keep needed context while removing or transforming completed answers through an explicitly supported source-derived operation.

\item \textsf{omit}: the solver does not need this anchor. An omitted artifact cannot impose a requirement absent from the request and remaining visible material.

\end{wbpromptitems}

Choose a supported operation for each \textsf{initial\_\allowbreak{}files} entry:

\begin{wbpromptitems}

\item \textsf{corpus\_\allowbreak{}file}: copy one selected file via \textsf{corpus\_\allowbreak{}ref.\allowbreak{}file\_\allowbreak{}id}.

\item \textsf{corpus\_\allowbreak{}dir}: copy one whole selected repository/archive tree. One entry preserves the tree; \textsf{focus\_\allowbreak{}paths} does not restrict the copy.

\item \textsf{corpus\_\allowbreak{}derived}: create a text file from identified original sources; describe the specific transformation. This cannot sanitize a binary or tree.

\item \textsf{llm\_\allowbreak{}direct}: generate a necessary small auxiliary text file.

\item \textsf{local\_\allowbreak{}tool}: generate a necessary binary or structured fixture using Python; it cannot assume access to corpus files that the runner has not supplied.

\end{wbpromptitems}

Do not retrieve additional material. Keep descriptions concise and limited to starting contents and required transformations. Do not add walkthroughs, diagnostic notes, field mappings, precomputed results, or solution checklists to the environment. Legitimate reference documentation can remain.

\textsf{setup\_\allowbreak{}steps} is an inventory of installable dependencies, not a place for code edits, task-solving commands, debugging advice, or conditional solution recipes. Do not assume the synthesis machine's installed packages exist for the solver. Dependencies and possible fallbacks do not prescribe a solver implementation. List dependency names or install commands without copying unverified version pins from the source project into mandatory setup. Keep construction reasoning in \textsf{notes}; do not claim a blueprint was executed.

\wbpromptheading{Evaluation criteria}

Write concise, checkable criteria for the instruction and the material the solver will actually receive. Criteria may be more detailed than the request, but every required or prohibited condition must follow from an explicit user requirement, an applicable visible specification or compatibility requirement, or a necessary condition for the result to be correct and usable.

Assess completion, correctness, grounding, and requested quality through final artifacts, behavior, or execution evidence. A preferred approach is not a correctness requirement. Existing implementation choices, internal plan notes, blueprint descriptions, and setup suggestions cannot create grading constraints. A quantity that can be computed, or an item shown in a reference, is not automatically a mandatory output.

Preserve valid alternatives. Keep required output paths and interfaces without prohibiting supporting files, tests, configuration, or internal refactoring. Do not require a library, technique, metric selection, or count unless the visible task contract actually requires it. Suggested verification methods are ways a reviewer can check the result, not mandatory steps for the solver. Accept equivalent results without requiring proof that a preferred library, tool, or rendering method was unavailable. Do not attach an availability condition to an otherwise valid alternative. State how to verify source-grounded results without precomputing them here.

Before returning, reconcile the three parts within this response: the environment supports the instruction, paths agree, and the rubric checks the stated work without adding hidden requirements. If another approach would satisfy the request and its applicable references, the rubric must accept it. Remove an unsupported restriction rather than expanding the instruction to justify it.

\wbpromptheading{Reference requests}

The following requests are copied from the original task collection. They illustrate only the instruction fields and their information density; they are not complete joint outputs or scenarios to imitate. Do not reuse their domains, paths, quantities, or requirements in a new task. Use the full current task definition below. Source records are listed in \textsf{docs/\allowbreak{}instruction\_\allowbreak{}few\_\allowbreak{}shots.\allowbreak{}json}.

\wbpromptheading{Reference request 1}

\wbpromptomission{[JSON example omitted]}

\wbpromptheading{Reference request 2}

\wbpromptomission{[JSON example omitted]}

\wbpromptheading{Reference request 3}

\wbpromptomission{[JSON example omitted]}

\wbpromptheading{Validated transition}

The task definition and persona below govern the requested work. Inspection notes and anchor analysis are internal construction context, not additional requirements to expose in the instruction or impose through the rubric.

\wbpromptvariable{\{transition\}}

\wbpromptheading{Language}

\wbpromptvariable{\{language\}}

\wbpromptheading{Workspace}

\wbpromptvariable{\{workspace\}}

\begin{samepage}

\wbpromptheading{Output}

Return one JSON object matching the schema, with no surrounding commentary. Use \textsf{status: "ready"} for a coherent joint specification. If the supported file operations cannot express an essential environment requirement, return \textsf{status: "unsupported"} and explain the limitation in \textsf{notes}.

\wbpromptomission{[JSON schema omitted]}

\end{samepage}
\end{wbpromptbox}
\nopagebreak[4]
{\captionsetup{type=figure,position=bottom,skip=6pt}\caption{Prompt template for task instantiation.}\label{fig:prompt_5}}

\end{document}